\documentclass[10pt,journal,compsoc]{IEEEtran}
\ifCLASSOPTIONcompsoc
  \usepackage[nocompress]{cite}
\else
  \usepackage{cite}
\fi
\ifCLASSINFOpdf
\else
\fi
\usepackage{epsfig}
\usepackage{amsmath}
\usepackage{amssymb}
\usepackage{pifont}

\usepackage{fancyhdr}
\usepackage{pdfpages}

\usepackage{blindtext}
\usepackage{booktabs, arydshln}
\usepackage{makecell}
\usepackage{multirow}
\usepackage{verlab}
\usepackage{xcolor}
\usepackage[utf8]{inputenc}

\DeclareMathOperator*{\argmax}{arg\,max}

\definecolor{darkgreen}{rgb}{0.0, 0.75, 0.0}
\definecolor{darkred}{rgb}{0.75, 0.0, 0.0}
\newcommand{\cmark}{\textcolor{darkgreen}{\ding{51}}}%
\newcommand{\xmark}{\textcolor{darkred}{\ding{55}}}%
\newcommand{\orcid}[1]{\hspace{1px}\raisebox{2px}{\href{https://orcid.org/#1}{\includegraphics[scale=0.04]{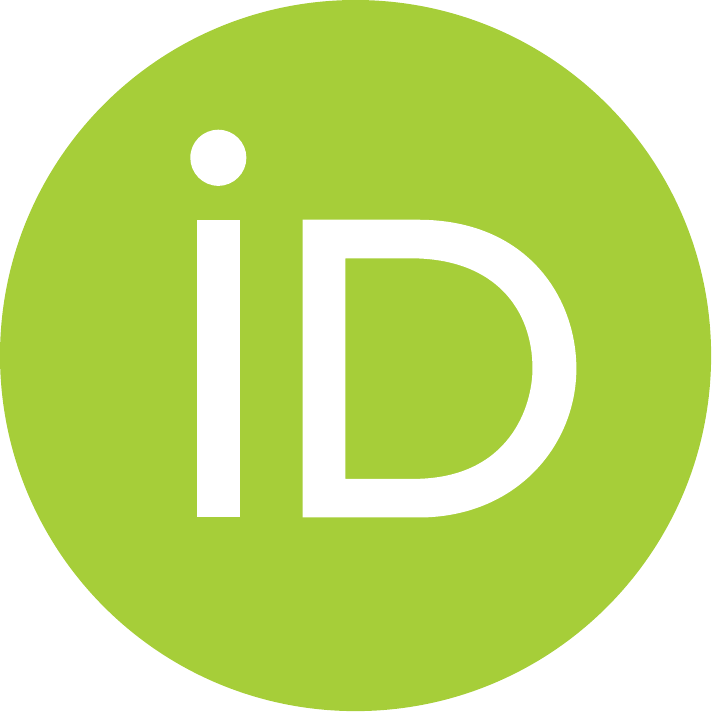}}}}

\usepackage{hyperref}

\usepackage{soul}

\makeatletter
\def\adl@drawiv#1#2#3{%
  \hskip.5\tabcolsep
  \xleaders#3{#2.5\@tempdimb #1{1}#2.5\@tempdimb}%
  #2\z@ plus1fil minus1fil\relax
  \hskip.5\tabcolsep}
\newcommand{\cdashlinelr}[1]{%
  \noalign{\vskip\aboverulesep
    \global\let\@dashdrawstore\adl@draw
    \global\let\adl@draw\adl@drawiv}
  \cdashline{#1}
  \noalign{\global\let\adl@draw\@dashdrawstore
    \vskip\belowrulesep}}
\makeatother

\begin{document}
%

\title{Text-Driven Video Acceleration: A Weakly-\\Supervised Reinforcement Learning Method}

%
%
%
%

\author{Washington~Ramos\orcid{0000-0002-0411-8677},
        Michel~Silva\orcid{0000-0002-2499-9619}
        Edson~Araujo\orcid{0000-0003-0585-5473},
        Victor~Moura\orcid{0000-0001-5379-8755},
        Keller~Oliveira\orcid{0000-0002-1287-7729},
        Leandro~Soriano~Marcolino\orcid{0000-0002-3337-8611},
        and~Erickson~R.~Nascimento\orcid{0000-0003-2973-2232}
\IEEEcompsocitemizethanks{\IEEEcompsocthanksitem Washington Ramos, Edson Araujo,
	Victor Moura, Keller Oliveira and Erickson R. Nascimento are with the Computer Science Deparment, Universidade Federal de Minas Gerais, Belo Horizonte, MG 31270-901, Brazil. E-mail: \{washington.ramos, edsonroteia, victorhugomoura, kellermartins, erickson\}@dcc.ufmg.br
	\IEEEcompsocthanksitem Michel Silva is with the Department of Informatics, Universidade Federal de Viçosa, Viçosa, MG 36570-900, Brazil. E-mail: michel.m.silva@ufv.br
	\IEEEcompsocthanksitem Leandro Soriano Marcolino is with Lancaster University, Lancaster, LA1 4YW, U.K. E-mail: l.marcolino@lancaster.ac.uk}
}

%
%

\markboth{Transactions on Pattern Analysis and Machine Intelligence}%
{Ramos \MakeLowercase{\textit{et al.}}: Text-Driven Video Acceleration: A Weakly-Supervised Reinforcement Learning Method}
%



\IEEEtitleabstractindextext{%
\begin{abstract}
The growth of videos in our digital age and the users' limited time raise the demand for processing untrimmed videos to produce shorter versions conveying the same information. Despite the remarkable progress that summarization methods have made, most of them can only select a few frames or skims, creating visual gaps and breaking the video context. This paper presents a novel \textit{weakly-supervised} methodology based on a reinforcement learning formulation to accelerate instructional videos using text. A novel joint reward function guides our agent to select which frames to remove and reduce the input video to a target length without creating gaps in the final video. We also propose the Extended Visually-guided Document Attention Network (VDAN+), which can generate a highly discriminative embedding space to represent both textual and visual data. Our experiments show that our method achieves the best performance in Precision, Recall, and F1 Score against the baselines while effectively controlling the video's output length. Visit \url{https://www.verlab.dcc.ufmg.br/semantic-hyperlapse/tpami2022/} for code and extra results.
\end{abstract}

}

\maketitle

\thispagestyle{fancy}
\fancyhf{}
\chead{{In IEEE Transactions on Pattern Analysis and Machine Intelligence (TPAMI) 2022. \\ The final is available at \url{https://doi.org/10.1109/TPAMI.2022.3157198}}}

\IEEEdisplaynontitleabstractindextext

%
\IEEEpeerreviewmaketitle

\IEEEraisesectionheading{\section{Introduction}\label{sec:introduction}}

\IEEEPARstart{T}{he} development of new digital technologies, especially over the last two decades, has allowed humans to overcome physical barriers in exchanging information. With the emergence of several tools such as portable devices (\eg, smartphones and wearables), and social multimedia services like Twitter, Facebook, Instagram, and YouTube, storing and sharing data of any kind has become effortless. Moreover, due to the exponential growth in processing power, storage, and bandwidth of digital devices, we are witnessing a substantial increase in the volume of data, in particular, textual and visual data such as images and videos. Every day a plethora of textual tutorials and instructional videos is published on the Internet teaching a variety of tasks, from how to cook burritos and tacos all the way to how to solve partial differential equations (PDEs).

Despite many textual tutorials and instructional videos sharing the increasing growth of available data as well as their content, they differ in a key aspect for users: {\it how long they would take to consume the content}. In general, information encoded by producers is more concise in the textual domain than in the visual domain. For instance, a recipe of tacos or a tutorial explaining how to solve a PDE is described in a few sentences. Instructional videos, for their turn, might have several minutes showing non-relevant information for the task such as a person opening a refrigerator, picking up the pencil, or erasing the blackboard. Such segments could be fast-forwarded without losing the message to be conveyed by the input video. Thus, ideally, instructional videos should be concise, similar to a textual description, but still having visually-rich demonstrations of all main steps of the task.
 
In this paper, we address the problem of accelerating temporally untrimmed videos by using text documents. For example, a recipe for cooking tacos could be used as a guide to select relevant frames from videos of cooking tacos. Note that this problem is different from moment localization~\cite{Hendricks2017Localizing}, procedure segmentation~\cite{Zhou2018Towards}, and video summarization~\cite{Plummer2017, Jung2020Global}, since non-relevant frames are still necessary for a user to understand the flow and temporal coherence of a task, \ie, some segments should be accelerated, but not eliminated.

Although identifying and fast-forwarding non-relevant segments may be a trivial task for humans, it poses significant challenges to fully automated systems. First, identifying relevant segments requires reasoning about the semantic concepts in the scene (\eg, person, tools, ingredients) and their interactions as they evolve over time. Second, usually in fast-forwarding techniques~\cite{Joshi2015, Halperin2017}, the overall length of the output video is defined by the user. Therefore, the method must adjust a varying speed-up rate along the footage to attend to the user's needs.

Over the past few years, algorithms on semantic fast-forwarding have been emerging as effective approaches to tackle the task of retrieving meaningful segments without losing the temporal continuity~\cite{Okamoto2013, Ramos2016, Lai2017, Lan2018, Silva2018, Silva2018cvpr, Ramos2020wacv, Ramos2020cvpr, Silva2021}. In our previous work~\cite{Ramos2020cvpr}, we initially introduced an algorithm to create fast-forward videos guided by text documents. We formulated our fast-forwarding task as a sequential decision-making process. A reinforcement learning (RL) agent locally decides to increase, decrease, or keep the video's speed-up rate based on the encoded text and video frame. An embedding space is generated by a Visually-guided Document Attention Network (VDAN), which creates representative feature vectors for textual and visual modalities. Although this RL agent and VDAN presented relevant results regarding the output content's relevance, they hold two significant drawbacks. First, the user does not control the output video's duration, which is one of the key aspects of most fast-forward algorithms. This control is also essential to systems where the storage or time resources are limited. Second, VDAN can only encode static frames; however, many sentences in the input document, such as the ones with actions describing an instruction step, may require modeling the scene temporal dynamics.

To overcome these drawbacks, we introduce the Skip-Aware Fast-Forwarding Agent (SAFFA), which can adapt its actions to drop frames according to their relevance and subject to a target speed-up rate defined by the user. SAFFA is trained with a novel joint reward function that aims at \textit{semantics} and \textit{speed-up}, allowing us to create videos of different lengths without retraining. We also present an extension to the representation capabilities of VDAN with the novel Extended Visually-guided Document Attention Network (VDAN+) that creates a joint embedding space for documents and video segments instead of simple static frames. Finally, thorough experiments are conducted on challenging instructional video datasets~\cite{Zhou2018Towards, Tang2020}. We show quantitative and qualitative results with in-depth discussions, showing that our method achieves the best performance in F1 Score against the baselines while effectively controlling the video's output length. 

This work takes a step forward towards fast-forwarding videos to a required length based on the video content's semantics and the speed-up rate defined by the user. Using textual data to guide an agent that seeks the best set of frames to be removed, our method emphasizes highly semantic content segments while preserving the temporal continuity and the target output length. Our approach is weakly supervised since it does not require any temporal annotation at training or inference time, \ie, only the textual instructions are needed. The problem of text-driven video acceleration naturally fits to the sequential decision-making formulation via a Markov Decision Process (MDP) since no information about previous states is required for taking subsequent actions. Unlike planning solutions like dynamic programming, which would require to go through all possible states multiple times to calculate the value function, our RL-based solution processes each state at most once. Thus, a trained agent would act similar to a knowledgeable human facing a new document and unseen video segments and controlling the video playback speed to achieve the required length.

Our contributions can be summarized as follows: \textit{i)}~a new fast-forward method based on a reinforcement learning formulation, which is able to accelerate videos according to clip similarity scores with textual data subject to a target speed-up rate; \textit{ii)}~a new state representation and joint reward function that guide our agent to attend the multiple goals, \ie, \textit{semantics} and \textit{speed-up}; \textit{iii)}~an extended version of the VDAN, the VDAN+, which can create a joint embedding space for documents and video clips, instead of static frames and; \textit{iv)}~extensive experiments with in-depth discussions, including an ablation study, using the YouCook2 and COIN datasets as opposed to a small subset of YouCook2 used in our previous work.
\section{Related Work}
\label{sec:related_work}

\subsection{Video Summarization}

Over the past several years, video summarization methods figured prominently in the task of shortening a video~\cite{Lee2012, Fei2017, Mahasseni2017, Zhang2016}. The shorter version is, usually, a summary of the input video composed of a storyboard of keyframes or video skims with the most distinguishable segments~\cite{Molino2017}. 

The strategy adopted to summarize the videos varies from frames features clustering~\cite{Mahasseni2017} and training neural networks that infer the representativeness of a video segment~\cite{Yao2016, Zhang2016} to employing additional information such as user queries, external sensors~\cite{Lee2012}, or textual annotations~\cite{Plummer2017}. Lee~\etal~\cite{Lee2012}, in the context of first-person videos, analyzed properties such as social interaction, gaze, and object detection to create a storyboard summary of the video. Zhang~\etal~\cite{Zhang2016} proposed a method to create a storyboard or skims by modeling long-range dependencies among the video frames using a Bi-LSTM network. Yao~\etal~\cite{Yao2016} performed the selection of the relevant segments fusing information from spatial and temporal CNNs to identify highlighting moments in videos of sports. Plummer~\etal~\cite{Plummer2017} created an approach that selects the best subset of video segments by analyzing their visual features along with vision-language modeling.

Reinforcement Learning (RL) has also been applied to video summarization~\cite{Zhou2018Deep, Li2021}, motivated by its success in many challenging tasks, such as mastering complex games like Go~\cite{Silver2017}, and achieving human-level performance in Atari~\cite{Mnih2015}. Additionally, it has great application in vision tasks, \eg, visual tracking~\cite{Yun2017} and image cropping~\cite{Li2018}. Zhou~\etal~\cite{Zhou2018Deep} presented an end-to-end unsupervised framework also based on RL. Their method summarizes videos by applying a diversity-representativeness reward that guides the agent to create more diverse and representative summaries. Li~\etal~\cite{Li2021} introduced an additional video level semantic reward to guide the unsupervised RL procedure proposed by Zhou~\etal. Their approach uses a semantic supervisor network to provide the reward signals based on the similarity between the original video and the generated summary. As a drawback, their video-level semantic reward may favor summaries that yield higher global semantic similarities, which is not the case of instructional videos where the steps may represent a small portion of the video. On the other hand, our approach provides local rewards given by the similarity between video segments and text. 

Despite impressive advances in shortening videos while retaining relevant segments, most of the summarization methods ignore temporal aspects or use a relaxed temporal restriction, resulting in visual gaps and breaking the video context.

\subsection{Semantic Fast-forwarding}

The lack of context that emerges from the gaps generated by summarization techniques creates a nuisance to video consumers, particularly instructional video consumers. The existence of a gap might confuse the user about the whole process. For instance, the user would be unaware if an important step was missed with the gap if the original video is unknown. Semantic fast-forward based methods add time constraint in the frame sampling, which results in a shorter and contiguous version of the input video. They usually apply different speed-up rates, where lower ones are assigned to the relevant segments and higher ones to the others.

Okamoto and Yanai~\cite{Okamoto2013} proposed a method to accelerate guidance videos emphasizing video segments with pedestrian crosswalks or turning movements in street corners. Ramos~\etal~\cite{Ramos2016} presented a semantic fast-forwarding method for first-person videos dealing with visual stability constraints with emphasis on human faces. Silva~\etal~\cite{Silva2018} extended the work of Ramos~\etal, including a CNN to assign frame scores based on Internet users' preferences and a video stabilizer proper to fast-forwarded videos~\cite{Silva2016}. Silva~\etal~\cite{Silva2018cvpr} proposed modeling the frame sampling in semantic fast-forwarding as a Minimum Sparse Reconstruction problem. The authors also propose an extension~\cite{Silva2021} that aims to remove visual gaps that could break the continuity of the output video, and to smooth the speed-up transitions between video segments. A drawback of the aforementioned works consists of pre-processing steps like detecting objects and computing optical flow, which is time-consuming and rely on the accuracy of third-party methods.

Lan~\etal~\cite{Lan2018} introduced the Fast Forward Network (FFNet). The network can summarize videos on the fly using an RL agent that selects frames with the most memorable views according to human-labeled data. Similar to FFNet and Zhou~\etal, we also apply an agent trained by the RL paradigm; however, our approach is a step towards training agents to work in a cross-modal embedding space in a weakly-supervised manner. Unlike FFNet, our approach does not demand human-labeled data to indicate the relevant frames at training time.

Most recently, methods in semantic fast-forwarding introduced usage of language to guide the frame selection. The approach proposed by Ramos~\etal~\cite{Ramos2020wacv} personalizes the fast-forwarding process by using social network posts as textual input. Their method uses a dense captioning algorithm on image frames to match positive tweets from Twitter in the embedding space. Although achieving a high F1 Score, the whole approach is too sensitive to errors from each of its components. 

\subsection{Cross-modal Embedding for Instructional Videos}

Recently, the algorithms on cross-modal embedding have emerged as promising and effective approaches to deal with tasks like video description~\cite{Pan2016} and text-based image/video retrieval~\cite{Mithun2019, Dong2018, Aytar2017}. Virtually, these methods rely on creating a shared embedding space, where features from multiple modalities can be compared.

There is a growing body research in instructional video and image analysis. Marin~\etal~\cite{Marin2019} applied a multi-modal neural model to learn a shared embedding space for images and recipes and tackled the task of retrieving recipes from image queries. Wang~\etal~\cite{Wang2019} proposed an adversarial learning strategy to align both modalities. Weakly supervised and unsupervised methods have been proposed to tackle different tasks such as activity segmentation~\cite{Sener2018},  localizing key steps~\cite{Alayrac2016}, and parsing~\cite{Sener2015} instructional videos. Leveraging cross-modal information, Alayrac~\etal~\cite{Alayrac2016} proposed a method to discover the main steps of a task by linking visual and textual clustering tasks to connect the modalities via joint constraints. Zhukov~\etal~\cite{Zhukov2019} proposed a new representation for instructional video steps by learning a set of classifiers per-component (\eg, nouns and verbs). Sener~\etal~\cite{Sener2015} use visual and language atoms to create a multimodal representation to parse a video into semantic steps.

In our previous approach~\cite{Ramos2020cvpr}, we built an end-to-end trainable embedding space for text and image, which is further used by an RL agent to accelerate an input video. However, unlike most of the other fast-forward methods~\cite{Kopf2014,Poleg2015,Joshi2015,Ramos2016,Halperin2017,Silva2018,Wang2018Hyper,Silva2018cvpr,Silva2021}, the agent cannot optimize the output speed-up rate of the video, which is essential in several applications. In this work, we extend our previous approach~\cite{Ramos2020cvpr} by introducing the Skip-Aware Fast-Forward Agent (SAFFA) and an Extended Visually-guided Document Attention Network (VDAN+). While our new agent can jointly optimize the semantics and the speed-up objectives, the new cross-modal embedding space, VDAN+, provides the semantic distance between each snippet in the instructional video and the textual steps described in a document, \ie, the recipe.
\begin{figure*}[t!]
	\centering
	\includegraphics[width=\linewidth]{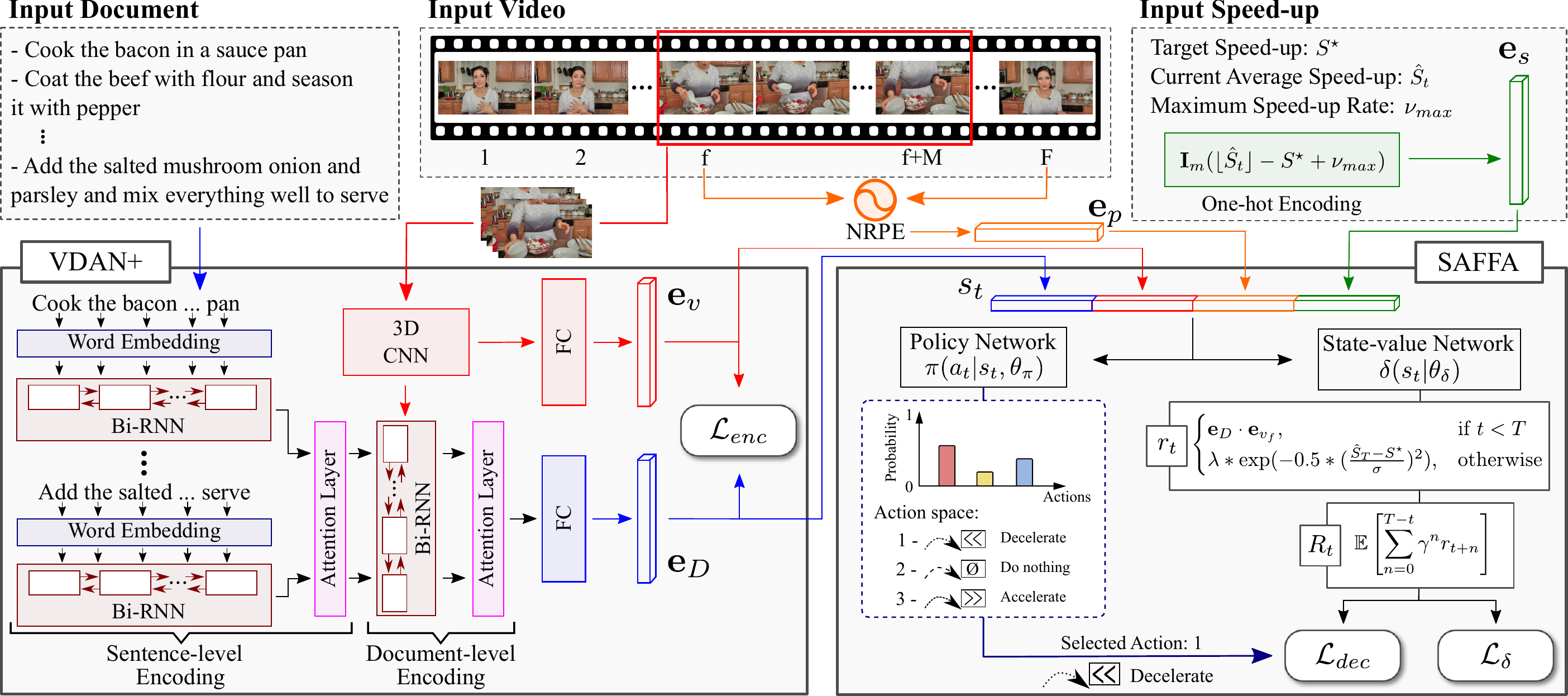}
	\caption{\textbf{Overview of our methodology.} It is composed of two main steps: \textit{i)} we employ our Extended Visually-guided Document Attention Network (VDAN+) to build a cross-modal embedding space that provides the representative embeddings $\mathbf{e}_D$ and $\mathbf{e}_v$ to the user document and the video segment, respectively; \textit{ii)} we train a reinforcement learning agent to select which frames to remove executing actions to increase, decrease, or keep the current skip rate given the embeddings $\mathbf{e}_D$ and $\mathbf{e}_v$, the encoded position in the video ($\mathbf{e}_p$), and the encoded average relative skip rate ($\mathbf{e}_s$).}
	\label{fig:methodology}
\end{figure*}

\section{Methodology}
\label{sec:methodology}

As stated, we formulate the fast-forwarding task as a sequential decision-making process, where an agent is trained to create fast-forwarded videos. Our methodology combines an RL paradigm and encode-decoding cross-modal framework to reduce videos' size by keeping the most relevant frames and avoiding temporal gaps. Specifically, after creating an embedding space for encoding documents and videos using a novel Extended Visually-guided Document Attention Network (VDAN+), we train an RL agent to decide which frames should be removed. The agent observes the encoded text and video snippets, its position in the video, and its current average skip rate, then outputs a distribution over the actions for increasing, decreasing, or maintaining the current speed-up rate. Figure~\ref{fig:methodology} shows the main steps of our approach.

\subsection{Extended Visually-guided Document Attention Network (VDAN+)}

In this paper, we extend VDAN~\cite{Ramos2020cvpr} with the novel Extended Visually-guided Document Attention Network (VDAN+). Although the embedding space created by VDAN has shown promising results providing semantically meaningful representation for image and text, it is limited by the lack of proper temporal modeling. For instance, some actions like ``open'' and ``close'' are inversely related; therefore, a single static frame would be ambiguous when associated with the text description. 

To overcome this limitation, VDAN+ takes a document and a video clip as input and, guided by the video features, creates representative feature vectors for both modalities. By training VDAN+, we aim at creating an embedding space in which textual and visual features are semantically aligned. We argue that the aligned embedding vectors help our agent make sense of the semantic proximity between the input document and the video frames and then learn the best policy to discard non-relevant frames as far as the document is concerned.

Formally, let ${D = \{p_1, p_2, \cdots, p_N\}}$ be a document composed of $N$ sentences, and $v$ a segment of length $M$ from the input video ${V = \{v_f\}_{f=1}^{F}}$ of $F$ frames. The VDAN+ produces $d$-dimensional embeddings ${\mathbf{e}_D \in \mathbb{R}^d}$ and ${\mathbf{e}_v \in \mathbb{R}^d}$ for textual and visual data, respectively. In our task, $D$ is represented by a document composed of a set of textual instructions.

\subsubsection{Document Encoder}

To encode $D$, we employ a Hierarchical Recurrent Neural Network (H-RNN) coupled with a soft attention mechanism in each level~\cite{Yang2016}. The usage of an H-RNN provides a lightweight architecture and aids the network in capturing long-range temporal dependencies by exploiting the nature of its hierarchically organized input data.

Our H-RNN comprises two levels of encoding: i) the sentence-level and ii) the document-level, as illustrated in Figure~\ref{fig:methodology}. Each level contains bi-directional GRU~\cite{cho2014learning} units that produce hidden state vectors that feed the attention layer. Let ${\mathbf{w}_{i1}, \cdots, \mathbf{w}_{iN_{i}}}$ denote the distributional word representations of each word in sentence $p_{i}$. The sentence-level encoder produces a hidden state vector ${\mathbf{h}_{ij} = f_p(\mathbf{w}_{ij}; \mathbf{h}_{i(j-1)}, \theta_{R_p})}$ at each timestep $j$ given the word embedding $\mathbf{w}_{ij}$, the previous hidden state $\mathbf{h}_{i(j-1)}$, and the parameters $\theta_{R_p}$. As stated by Yang~\etal~\cite{Yang2016}, words have different contributions to the meaning of a sentence. Therefore, we feed $\mathbf{h}_{ij}$ to the attention module defined as:
${\mathbf{u}_{ij} = \text{tanh}(W_p\mathbf{h}_{ij} + b_p)}$ and ${\alpha_{ij} = \text{exp}(\mathbf{u}_{ij}^\intercal c_p)/\sum_{j^\prime} \text{exp}(\mathbf{u}_{i{j^\prime}}^\intercal c_p)}$.
\noindent $W_p$ is a learnable projection matrix and $c_p$ is a learnable word-level context vector that acts as a fixed query to find the informative word. The alignment between $c_p$ and $\mathbf{u}_{ij}$ defines the score used to compute the weight $\alpha_{ij}$ that gives the importance for each $\mathbf{h}_{ij}$. ${\mathbf{p}_i = \sum_j \alpha_{ij}\mathbf{h}_{ij}}$ is the sentence-level embedding for the sentence $p_{i}$.

In the document-level encoding, each $\mathbf{p}_i$ is used to produce a hidden state vector ${\mathbf{h}_{i} = f_d(\mathbf{p}_{i}; \mathbf{h}_{i-1}, \theta_{R_d})}$. Different sentences may also contribute differently to the document. In our approach, the instructional characteristic of the document increases the probability of a given video segment being more similar to a few instructions rather than the whole document. Thus, similar to the sentence-level counterpart, we employ an attention module, which is parameterized by $W_d$ and $c_d$. As a result, after feeding the document-level encoder with all vectors $\mathbf{p}_{i}$, it yields the document-level encoding $\mathbf{d}$. Finally, we encode the document as ${\mathbf{e}_D = f_D(\mathbf{d}; \theta_D)}$, \ie, we project $\mathbf{d}$ into the embedding space using the fully connected network $f_D$ parameterized by $\theta_D$.

\subsubsection{Video Encoder}

To produce the clip embedding $\mathbf{e}_v$, we first extract the video features with a 3D Convolutional Neural Network, producing an intermediate vector ${\phi(v) \in \mathbb{R}^z}$. In this work, we use the features produced by the penultimate layer of the R(2+1)D-34~\cite{Tran2018R2plus1D} pretrained on $65$ million Instagram videos (IG-65M)~\cite{Ghadiyaram2019}. Then, we project $\phi(v)$ into the embedding space using a fully connected network $f_v$ parameterized by $\theta_v$ as follows ${\mathbf{e}_v = f_v (\phi(v); \theta_v )}.$ 

To facilitate training and guide the attention weights to the sentences that best characterize the input video, we condition the creation of the document embedding, $\mathbf{e}_D$, to the clip features $\phi(v)$. Specifically, we set the first hidden state vector of the document-level encoder as ${\mathbf{h}_0 = \phi(v)}$. This strategy adds context to our document encoder in a similar manner that image and video captioning approaches do to decode the captions~\cite{Vinyals2015, You2016}. It is noteworthy that although $D$ remains the same, $\mathbf{e}_D$ is not unique throughout the input video since $\phi(v)$ may change.

Both document and video encoding modules include an $\ell_2$ normalization layer to make $\mathbf{e}_D$ and $\mathbf{e}_v$ unit norm vectors. We found in our experimentation that using a batch normalization layer~\cite{Ioffe2015Batch} preceding the $\ell_2$ normalization layer increased the performance of our model.

\subsubsection{Training}
\label{sec:methodology_training}

We follow a pairwise training strategy to build the cross-modal embedding space. For each clip $v$ in the training set, we create a positive and a negative document, $D^+$ and $D^-$, to compose the training pairs $<$$D^+, v$$>$ and $<$$D^-, v$$>$. A positive document $D^+$ comprises sentences that describe the clip $v$ and, additionally, sentences describing a randomly selected clip $v^\prime$. The strategy of adding sentences that do not describe the clip assists the document-level attention module on attending the proper sentences at training time. To create the negative document, $D^-$, we randomly select two other clips $v^{\prime\prime}$ and $v^{\prime\prime\prime}$, and collect their respective sentences. At each training step, we shuffle all the sentences in the document for generalization purposes.

In order to create more aligned embeddings, we optimize ${\theta_{enc} = \{\theta_{R_{p}}, W_p, c_p, \theta_{R_{d}}, W_d, c_d, \theta_D, \theta_v\}}$ by minimizing the cosine embedding loss as follows:
\begin{equation}
	\small
    \mathcal{L}_{enc}(\mathbf{e}_D, \mathbf{e}_v; y) = 
    \begin{cases}
        1 - \cos(\mathbf{e}_D, \mathbf{e}_v), & \text{if}\ y = 1\\
        \max(0, \cos(\mathbf{e}_D, \mathbf{e}_v) - \eta), & \text{otherwise},
    \end{cases}
	\label{eq:lenc}
\end{equation}
\noindent where $y$ is equal to $1$ if $\mathbf{e}_D$ and $\mathbf{e}_v$ were generated via a positive pair, and $\eta$ is a margin parameter, which is set to $0$ in our problem.

\subsection{Skip-Aware Fast-Forward Agent (SAFFA)}

After building the cross-modal embedding space, we train an agent to observe the encoded vectors $\mathbf{e}_D$ and $\mathbf{e}_v$, its position in the input video, and its current average speed-up rate, then sample an action from the action space to adjust its speed-up rate accordingly.

We formulate the problem of selecting frames as a Markov Decision Process (MDP). In our formulation, we train an agent to maximize the expected sum of discounted rewards: ${R_t = \mathbb{E}\left[\sum_{n=0}^{T - t}\gamma^n r_{t+n}\right]}$, where $t$ is the current timestep, $r_{t+n}$ is the reward $n$ timesteps into the future, and $T$ is the total number of timesteps. At each timestep, one frame is selected; therefore, $t$ also indicates the current number of selected frames, and $T$ the total number of selected frames. ${\gamma \in (0,1]}$ is a discount factor.

In our problem, the agent observes a long input video and a text document, and takes actions to create an accelerated version of the video. Since we aim to keep the overall coherence of the video instead of trimming it, we restrict the agent to navigate the video space observing short segments and skipping them accordingly, limited to a maximum skip length. The agent has velocity $\nu$ and acceleration $\omega$, and based on the current $\nu$, the next frame is selected. Therefore, the agent goes through the whole video, but skips frames according to a dynamically changing velocity. At each timestep, the agent can increase, decrease, or keep its current acceleration, which will, in turn, affect the velocity. Since we apply Model-free Reinforcement Learning, the transition function does not need to be pre-defined, nor learned; as the agent focus directly on learning the best policy. In the following sections, we define all elements used in our MDP formulation.

\subsubsection{Action Space and States Composition} 

To use text documents as guide while accelerating a video, our agent adaptively adjusts the skip rate $\nu$ (velocity) to ensure lower speed to the video segments semantically similar to the input text and higher speed otherwise. The agent's action space $\mathcal{A}$ has three elements: i) \textit{decelerate}; ii) \textit{do nothing}; and iii) \textit{accelerate}. These actions are used by the agent to update the skip rate as follows: \textit{decelerate} and \textit{accelerate} update the velocity and acceleration states of the agent as ${\nu = \nu - \omega}$ and ${\omega = \omega - 1}$ for \textit{decelerate}, and ${\nu = \nu + \omega}$ and ${\omega = \omega + 1}$ for \textit{accelerate}, while \textit{do nothing} keeps the current $\nu$ and $\omega$. Acceleration and velocity saturate at values empirically set as  ${\omega_{max}=5}$ and ${\nu_{max}=25}$, respectively, which are always greater than or equal to $1$. Note that $ \omega $ does not correspond to a physical acceleration, allowing the agent to quickly adjust the velocity to collect more rewards when the semantic level changes or to focus on the target speed-up rate.

To allow our agent to navigate through the video space attending to the requirements effectively, the state needs to encode information about the current semantics and the agent's location in the video. Thus, the agent would be able to reason whether the skip rate should be higher or lower, aiming to create an output with length as close as possible to the desired one while maximizing the exhibition of scenes related to the document. A straightforward state composition to achieve these requirements would be concatenating the document's features, the current video segment, and the agent's average skip rate. However, by observing such a state, the agent would be unaware of its distance to the end of the video, while a successful decision to increase or decrease the skip rate depends on it. For example, at the beginning of the video, the agent has more freedom to attend to both objectives. However, as the agent approaches the end of the video, attending the target speed-up rate becomes more challenging since changes in the current skip rate present a negligible effect on the average skip rate. Therefore, inspired by the successful usage of positional encoding in recent works~\cite{Vaswani2017Attention, Takase2019Positional, Jung2020Global}, we propose the Normalized Reversed Positional Encoding (NRPE) to encode the current agent's location in the video.

Let $q$ be the NRPE embedding size, $F$ be the number of frames in the input video and $f$ the position of the agent in the video, \ie, ${f \in \{1, 2, \cdots, F\}}$. Then, the dimensions ${2k}$ and ${2k+1}$ (with ${k \in \{1, 2, \cdots, q/2\}}$) of our NRPE embedding, $\mathbf{e}_p$, are:
\begin{equation}
	\footnotesize
	\label{eq:nrpe}
	\begin{aligned}
		NRPE_{(f,2k)} = \sin\Bigg (\frac{F - f}{F^{\frac{2k}{q}}}\Bigg ),\;		
		NRPE_{(f,2k+1)} = \cos\Bigg (\frac{F - f}{F^{\frac{2k}{q}}}\Bigg ).
	\end{aligned}
\end{equation}
The rationale of NRPE is that an agent in videos of different lengths but in the same relative position and under the same relevance profile should behave the same.

To encode the agent's average skip rate, we use the one-hot vector given by ${\mathbf{e}_s = \mathbf{I}_m(\lfloor\hat{S}_t\rfloor - S^\ast + \nu_{max})}$, where $ \hat{S}_t $ is the average skip rate at the timestep $t$, $ {S^\ast \le \nu_{max} \in \mathbb{N}^+} $ is the target speed-up rate, and $ {\mathbf{I}_m(\iota)} $ denotes the $ \iota^{th} $ line of an identity matrix of size $ m $, set as ${m=2\nu_{max}}$ in our experiments. We use the average relative skip rate, ${\lfloor\hat{S}_t\rfloor - S^\ast}$, instead of the raw average skip, $\hat{S}_t$, to allow the agent to decide to increase or decrease its speed according to the deviation to the target speed-up rate. Hence, a single agent can accelerate an input video at different speed-up rates with no need to be re-trained. The final state composition at a given timestep $ t $ is defined as: ${s_t = [\mathbf{e}_D;\mathbf{e}_v;\mathbf{e}_p;\mathbf{e}_s]^\intercal \in \mathcal{S}}$.

\subsubsection{Reward function}

The goal of our agent is to learn a policy ${\pi(a | s_t, \theta_\pi)}$ that represents the probability of taking a certain action ${a \in \mathcal{A}}$ given the state $s_t$ and the parameters $\theta_\pi$. The reward should encourage the agent to increase, decrease, or keep its skip rate \wrt the semantic similarity between the textual and visual data in the upcoming video segment while considering the overall speed-up rate objective in the long-term. Therefore, we design an immediate reward proportional to the alignment of the textual and visual features in non-terminal states and proportional to the overall speed-up rate deviation at the terminal state (final frame of a video). Thus, at training time, after taking the action ${a_t \sim \pi(a | s_t, \theta_\pi)}$ in the $t^{th}$ step, the agent receives the following reward signal:
\begin{equation}
	\label{eq:immediate_reward}
	r_t = 
	\begin{cases}
		\mathbf{e}_D \cdot \mathbf{e}_v, & \text{if}\ t < T\\
		\lambda * \exp({-0.5 * (\frac{\hat{S}_T - S^\ast}{\sigma})^2}), & \text{otherwise},
	\end{cases}	
\end{equation}
\noindent where $\lambda$ controls the relative importance of the overall speed-up rate \wrt the frames' relevance in the output video. Note that the terminal reward is similar to a Gaussian function centered at $S^\ast$. Thus, lower ${\sigma \in \mathbb{R}}$ values force the agent to achieve the desired skip more accurately. Semantically, the agent receives higher rewards if $\mathbf{e}_D$ and $\mathbf{e}_v$ point to the same direction in the embedding space, which encourages the agent to reduce the speed and accumulate more rewards since the neighboring temporal frames are more likely to yield higher reward values.

Recall that our final objective is to maximize the expected sum of discounted rewards. Therefore, although being established as a terminal reward, the speed-up rate signal may, potentially, affect all the agent's decisions.

\subsubsection{Policy Learning}

Apart from aligning the textual and visual features produced by VDAN+, the overall objective of our methodology also tries to maximize the expected cumulative reward $R_t$ at each timestep $t$. We follow the REINFORCE algorithm~\cite{Williams1992} to learn the parameters $\theta_\pi$ that maximizes the expected utility: ${J(\theta_\pi) = \sum_{a \in \mathcal{A}} \pi(a | s_t, \theta_\pi) R_t}$.

To improve learning performance, we use the advantage function approach~\cite{Sutton2018} and maximize the expected advantage:
\begin{equation}
J^\prime (\theta_\pi) = \sum_{a \in \mathcal{A}} \pi(a | s_t, \theta_\pi)(R_t - v(s_t | \theta_v)),     
\end{equation}
\noindent where $v (s_t | \theta_v)$ is a function parameterized by $\theta_v$, which predicts our expected cumulative reward at state $s_t$. The gradient of $J^\prime$, $\nabla_{\theta_\pi}J^\prime(\theta_\pi)$, is given by:
\begin{equation}
 \sum_{a \in \mathcal{A}} \pi(a | s_t, \theta_\pi) (\nabla_{\theta_\pi} \log \pi(a | s_t, \theta_\pi)) (R_t - v(s_t|\theta_v)). 
\end{equation}
Usually, Monte Carlo sampling is applied, due to the high dimension of the action sequence space, leading to the following approximation for the gradient: 
\begin{equation}
\nabla J^\prime(\theta_\pi) \approx \sum_t \nabla_{\theta_\pi} \log \pi (a_t | s_t, \theta_\pi) (R_t - v(s_t|\theta_v)),
\end{equation}
\noindent where $a_t$ is the action taken at time $t$. Hence, we minimize the following loss function:
\begin{equation}
\mathcal{L}^\prime (\theta_\pi) = - \sum_t \left(\log \pi(a_t | s_t, \theta_\pi)\right) (R_t - v(s_t | \theta_v)).
\end{equation}
Moreover, in order to have a greater action diversity, we add the entropy of the policy output $H(\pi(a_t | s_t,\theta_\pi))$ into the loss~\cite{Li2018}. Therefore, our final policy loss is given by
\begin{equation}
\mathcal{L}_{dec} (\theta_\pi) = \mathcal{L}^\prime(\theta_\pi) - \sum_t \beta \cdot H(\pi(a_t|s_t,\theta_\pi)),
\label{eq:ldec}
\end{equation}
\noindent where $\beta$ is a constant to balance the entropy importance. In our experiments, we set $\beta$ to $0.01$.

Additionally, we also need to learn the state-value function $\delta(s_t | \theta_\delta)$. We do that by minimizing the mean squared error: ${\mathcal{L}_\delta (\theta_\delta) = \sum_t \left(\delta(s_t | \theta_\delta) - R_t \right)^2}$. Both losses $\mathcal{L}_\delta$ and $\mathcal{L}_{dec}$ can now be minimized using stochastic gradient descent. At test time, we use $ \argmax_a \pi(a|s_t,\theta_\pi)$ as the chosen action for the agent in a given timestep $t$.

\begin{table*}[t!]
	\centering
	\caption{\textbf{Comparison with baselines.} Precision, Recall, F1 score, Output Speed-up (OS), and Overall Performance (OP) in YouCook2 and COIN datasets. The estimated running times are on the right, and the target OS values are ${S^\ast=12}$ for YouCook2 and ${S^\ast=16}$ for COIN. Best values are in bold.}
	\label{tab:main_results}
	\setlength{\tabcolsep}{1.5pt}
	\scriptsize{
		\begin{tabular}{lccccccccccccccccccccccccccccccccc}
			\toprule
			\multicolumn{1}{c}{\multirow{2}{*}{\thead{\scriptsize{\textbf{Method}}}}} & & \multirow{2}{*}{\thead{\scriptsize{\textbf{No Dense}}\\\scriptsize{\textbf{Supervision}}}} & & \multirow{2}{*}{\thead{\scriptsize{\textbf{Controls}}\\\scriptsize{\textbf{Speed-up}}}} & & \multirow{2}{*}{\thead{\scriptsize{\textbf{Uses}}\\\scriptsize{\textbf{Language}}}}  
			& & & \multicolumn{9}{c}{\thead{\scriptsize{\textbf{YouCook2}}}} & & & \multicolumn{9}{c}{\thead{\scriptsize{\textbf{COIN}}}} & & & \multirow{2}{*}{\thead{\scriptsize{\textbf{Running}}\\\scriptsize{\textbf{Time (ms)$^3$}}}} \\
			& & & & & & & & & Precision$^1$ & & Recall$^1$ & & F1 Score$^1$ & & OS$^2$ & & OP$^1$ & & & Precision$^1$ & & Recall$^1$ & & F1 Score$^1$ & & OS$^2$ & & OP$^1$	\\  
			\cmidrule{3-7} \cmidrule{10-18} \cmidrule{21-29} \cmidrule{32-32}
			FFNet \cite{Lan2018} 	    	& & \xmark  & & \xmark & & \xmark & & & $ 58.83 $ 			& & $ 11.70 $ 			& & $ 18.86 $ 		    & & $ 11.90 $ 			& & $ 31.71 $ 		 	& & & $ 45.06 $ 			& & $ 14.08 $           & & $ 17.66 $          	& & $ 16.45 $ 			& & $ 29.75 $ 			& & & $ 08.65 $ 			\\
			\cdashlinelr{1-32}
			SAS \cite{Silva2018cvpr} 		& & \cmark & & \cmark & & \xmark & & & $ 49.22 $ 			& & $ 08.61 $           & & $ 14.44 $          & & $ 11.64 $ 			& & $ 25.02 $         	& & & $ 40.17 $ 			& & $ 09.07 $			& & $ 13.20 $ 			& & $ 16.10 $ 			& & $ 23.32 $ 			& & & $ 344.15 $ 			\\
			SASv2 \cite{Silva2021} 			& & \cmark & & \cmark & & \xmark & & & $ 49.69 $ 			& & $ 09.87 $ 			& & $ 16.20 $          & & $ 10.32 $ 			& & $ 19.60 $ 		 	& & & $ 41.29 $ 			& & $ 09.79 $     		& & $ 13.90 $ 			& & $ 14.09 $ 			& & $ 20.07 $ 			& & & $ 344.43 $ 			\\
			BoT \cite{Ramos2020wacv} 	    & & \cmark & & \cmark & & \cmark & & & $ 48.66 $ 			& & $ 08.37 $ 			& & $ 14.04 $          & & $ \mathbf{12.13} $ 	& & $ 24.40 $ 		 	& & & $ 39.80 $ 			& & $ 09.21 $      		& & $ 13.01 $     		& & $ \mathbf{16.07} $	& & $ 23.02 $ 			& & & $ 1048.01 $ 			\\
			Ours 							& & \cmark & & \cmark & & \cmark & & & $ \mathbf{53.20} $ 	& & $ \mathbf{12.81} $ 	& & $ \mathbf{17.86} $ & & $ 11.68 $ 			& & $ \mathbf{30.07} $	& & & $ \mathbf{42.58} $ 	& & $ \mathbf{15.94} $	& & $ \mathbf{17.18} $	& & $ 14.99 $ 			& & $ \mathbf{27.98} $ 	& & & $ \mathbf{75.58} $ 	\\
			\cmidrule{10-32}
			&&&&&&&&&& \multicolumn{22}{r}{\scriptsize{\textit{$^1$Higher is better (\%)} \textit{$^2$Better closer to $S^\ast$}} \textit{$^3$Lower is better}} \\
			\bottomrule 
		\end{tabular}
	}
\end{table*}

\section{Experiments}
\label{sec:experiments}

\subsection{Datasets, Baselines, and Evaluation Metrics}

We conducted our experiments on the YouCook2 and COIN datasets~\cite{Zhou2018Towards, Tang2020}. YouCook2 is a large-scale dataset composed of unconstrained YouTube cooking videos, where the footages include a variety of cuisines and cooking styles. It consists of $2{,}000$ videos distributed across $89$ recipes with a total length of $176$ hours. COIN is a large-scale dataset composed of $11{,}827$ instructional videos distributed across $180$ tasks organized hierarchically into $12$ domains with a total length of $476$ hours. Videos in both datasets are annotated with temporal boundaries and natural language descriptions. Because YouCook2 lacks a test set with available textual instructions, we used its validation set as our test set. To optimize the hyperparameters and evaluate our model, we used a strategy commonly applied in the reinforcement learning literature (\eg, DQN~\cite{Mnih2015}, A3C~\cite{Mnih2016}, Rainbow~\cite{Hessel2018}, and DDQN~\cite{Hasselt2016}). First, we randomly selected a subset of the recipes (in our case, $10\%$) from the entire dataset and used their videos to tune the hyperparameters. Then, using all videos from the training and validation sets, including the ones used in the hyperparameter tuning, we train and evaluate our model.

To evaluate the performance of each method, we use the F1 Score (F1), which is the harmonic mean of the Precision and Recall. The higher the F1, the better retrieving relevant frames with the lowest false positive and negative rates. Note, however, that due to the constraints in temporal connections imposed to fast-forwarding techniques, they naturally select non-relevant frames since the relevant segments are usually distant from one another. We also measure the methods' capability of accelerating the input videos regarding the target speed-up rate. For that, we report the Output Speed-up (OS) rate, $\hat{S}$. The closer is $\hat{S}$ to $S^\ast$, the better.

There is a clear trade-off between F1 and OS. Selecting more frames may increase the F1 score while failing to attend the speed-up. Therefore, we propose the Overall Performance (OP) metric, which summarizes both F1 and OS in a single value. We compute the OP as the harmonic mean between the F1 and the OS accuracy, which is given by a Gaussian function centered at $S^\ast$, and with a standard deviation $\sigma_{OS}$. In the literature, acceptable speed-up rate errors are on average $5.31\%$, with a standard deviation of $8.38\%$ \wrt the target~\cite{Kopf2014, Joshi2015}. We use this standard deviation in our evaluation, \ie, ${\sigma_{OS}=0.0838\times S^\ast}$.

We compared our method with the following fast-forwarding techniques: Sparse Adaptive Sampling (SAS)~\cite{Silva2018cvpr} and its extension SASv2~\cite{Silva2021}, which are state of the art in terms of semantics retained in the final video; the Bag-of-Topics (BoT) technique~\cite{Ramos2020wacv}, which, like ours, also has texts and videos as input and; our previous approach~\cite{Ramos2020cvpr} with the semantic encoder VDAN and the Semantic Fast-Forward Reinforcement Learning (SFF-RL) agent. We also report results for the Fast-Forward Network (FFNet), an RL agent proposed by~Lan~\etal~\cite{Lan2018}. It is worth noting that FFNet is used as a gold standard for semantics since it disregards the output speed-up as a target. During experiments in the COIN dataset, the FFNet agent did not converge, acting as a uniform selection method due to the high variability of labels from different domains. For the sake of a fair comparison, we performed the experiments in COIN in a domain-wise manner.

\subsection{Implementation Details}
\label{sec:implementation_details}

We adopted a stage-wise training strategy, where VDAN+ and SAFFA are trained in two distinct stages. With this strategy, we avoid sending different reward signals to the agent during training and make the convergence of the model easier~\cite{Barshan2015}. Therefore, we first train VDAN+ using pairs of documents and clips from the VaTeX dataset~\cite{Wang2019VaTeX}. All the available English descriptions in the dataset are used to compose $D^+$ and $D^-$. We chop the sentences to be no longer than $20$ words and use a set of glove embeddings pre-trained in the \textit{Wikipedia 2014} and \textit{Gigaword 5} sets~\cite{Pennington2014} to represent each word. Note that, although a set of annotated descriptions for a given clip may not fairly represent an instructional document structure, the first training stage is mainly dedicated to creating a highly discriminative embedding space for aligning documents and video clips.

We tested the sizes $\{128,512,1{,}024\}$ for the dimension of the embedding space $d$, and we choose ${d = 128}$ as it showed a similar performance, but with fewer parameters. We define $512$ to be the size of the hidden state vectors $\mathbf{h}_{ij}$ and $\mathbf{h}_i$, and $1{,}024$ to be the size of $c_p$ and $c_d$. Note that, in our semantic encoder, the $\mathbf{h}_i$ vectors must lie on the same dimension as the visual features since ${\mathbf{h}_0=\phi(v)}$. The functions $f_v$ and $f_D$ are implemented as two independent fully connected neural networks composed of a single hidden layer with $512$ neurons.

\begin{figure}[t!]
	\centering
	\includegraphics[width=\columnwidth]{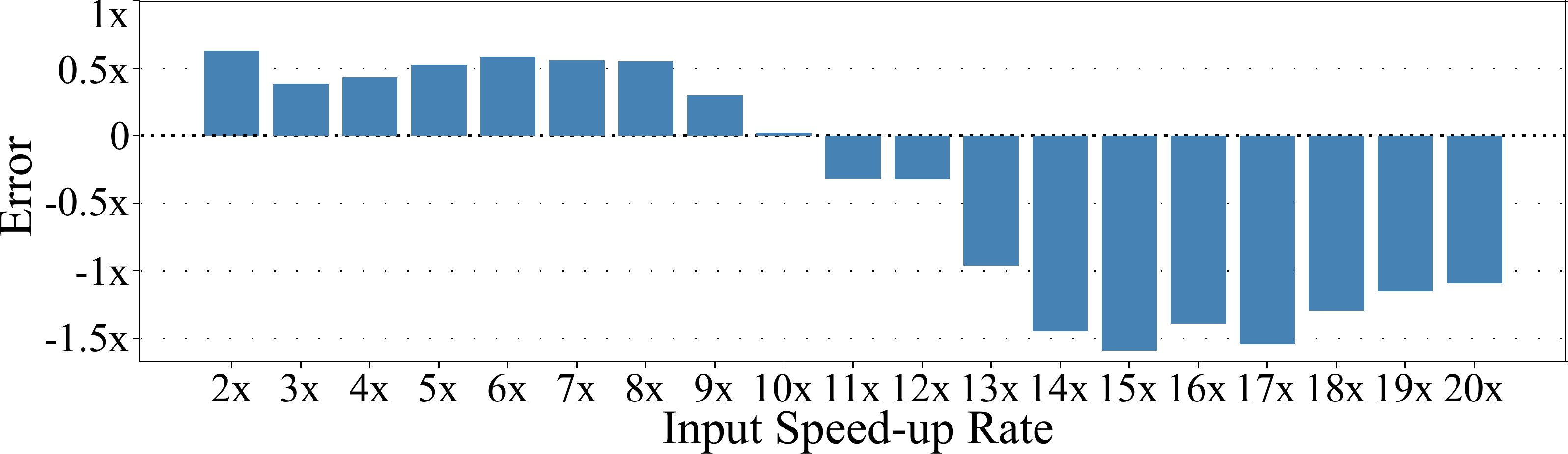}
	\caption{\textbf{Average output speed-up rate error in the YouCook2 dataset.} For target speed-up rates from $2$ to $20\times$, SAFFA generates videos with a small relative error, indicating its accuracy in the speed-up objective.}
	\label{fig:barplot_ablation_speedup_progress}
\end{figure}

We train VDAN+ for $100$ epochs with a batch size of $64$ and obtain the model that had the best performance in the validation set. During training, we rescale the videos to ${128\times171}$, then randomly cropped them spatially to ${112\times112}$~\cite{Ghadiyaram2019}. We also apply clip-wise horizontal random flip with $0.5$ probability and temporal jittering, obtaining clips of $32$ continuous frames.

The policy network ${\pi(a_t | s_t,\theta_\pi)}$ and the state-value function ${\delta(s_t | \theta_\delta)}$ were implemented as two Multilayer Perceptrons (MLPs) with two hidden layers ($256$ and $128$ neurons) each using ReLu activation function. We added a final layer with ${|\mathcal{A}|=3}$ neurons and softmax activation to ${\pi}$, and one with a single neuron and linear activation to the ${\delta}$ network.

\begin{figure*}[t!]
	\centering
	\includegraphics[width=\linewidth]{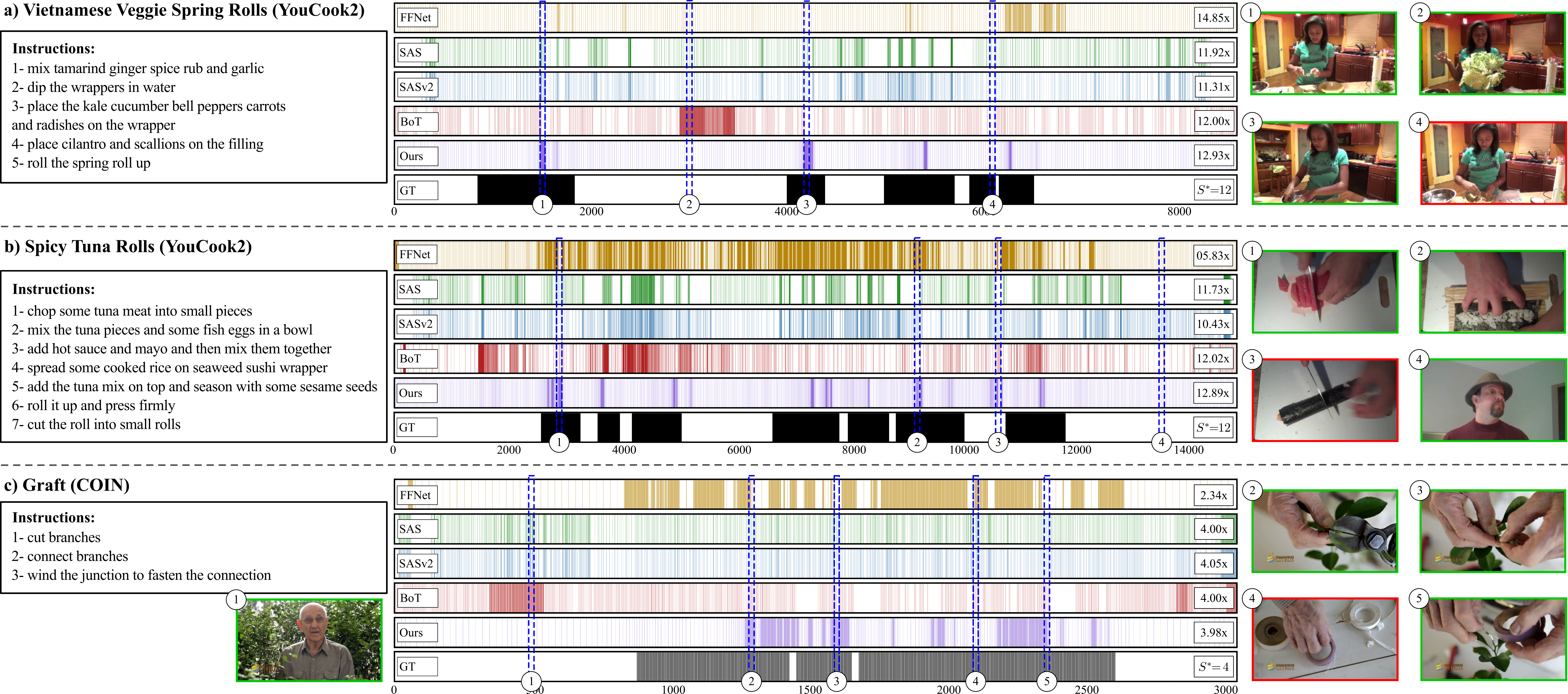}
	\caption{\textbf{Qualitative results.} The colored bars (right) represent the frames selected by each method for each instructional video, and the contiguous black blocks represent the ground-truth segments. Note that, in general, our agent performs a denser frame sampling in the regions representing a instruction step and a sparser one in the other regions (see the images outlined in green and the instructions). Exceptions are the regions where our agent decides to increase its skip rate to attend the target speed-up or matches the video snippet with the instructions (images outlined in red).}
	\label{fig:qualitative_evaluation}
\end{figure*}
We ran a grid search in the YouCook2 dataset to find the best policy learning rate $lr \in\{1\mathrm{e}{-5}, 5\mathrm{e}{-5}, 1\mathrm{e}{-4}\}$, $\sigma \in \{0.5, 1, 2\}$, and $\gamma \in \{0.8, 0.9, 0.99\}$, with ${\lambda=F^\ast}$, where $F^\ast$ is the desired number of frames. We selected the configuration with policy learning rate of $5\mathrm{e}{-5}$, $\sigma$ of $0.5$, and $\gamma$ of $0.99$, which provided the highest F1 Score such that $round(\hat{S})=S^\ast$. This configuration was fixed for all experiments. In our experiments, $\gamma$ and $\sigma$ have shown to be the most influential hyperparameters. Lower $\gamma$ values reduce the terminal reward's strength in the first input video frames, making it difficult to control the overall speed-up, and higher $\sigma$ values permit the agent to select more frames, reducing the overall speed-up without additional punishment. Random initialization did not influence the results.

We trained SAFFA and its variants for $100$ epochs using the Adam optimizer~\cite{Kingma2017}. In this stage, all VDAN+ parameters remain frozen. For faster convergence, we use a learning rate of $1\mathrm{e}{-3}$ for the value-state approximator. Our approach is fully implemented in the PyTorch library, and the experiments were conducted in an Intel\textsuperscript{\textregistered} Core\textsuperscript{\texttrademark} i7-3770K CPU @ 3.50GHz machine with 32GB RAM and an NVIDIA Titan RTX GPU.

\subsection{Results}

\subsubsection{Quantitative Results}

Table~\ref{tab:main_results} shows the average Precision, Recall, F1 Score~(F1), and Output Speed-up~(OS), as well as the Overall Performance~(OP), achieved by each method in both datasets. The methods are grouped according to whether they use labels for training (if applicable), optimize the video's output length, and use natural language as input. The gold standard method, FFNet, is above the dashed line as it needs frame-level (dense) labels for supervision. Because FFNet is the only method that does not allow a target speed-up rate as input, we used its average output speed-up rates, ${\hat{S}=11.90}$ and ${\hat{S}=16.45}$, as targets for all methods. \Ie, ${S^\ast=12}$ for the YouCook2 dataset and ${S^\ast=16}$ (on average) for the COIN dataset since ${S^\ast \in \mathbb{N}^+}$. We removed the COIN domains ``Nursing and Care'' and ``Leisure and Performance'' from the evaluation since ${S^\ast=1}$ and it would not require any acceleration for the other approaches. We also present the estimated running times in milliseconds (ms), where we can see that our approach is superior to the competitors. We refer the reader to the supplementary materials for detailed time analysis. The best values in Table~\ref{tab:main_results} are in bold.

The results indicate that our approach outperforms all competitors by a significant margin in Precision, Recall, and F1, while not compromising the output speed-up rate significantly in both datasets, which is reflected in the OP metric. It means that SAFFA could use the instruction steps described in natural language and match them to what is currently being observed in the scene. On the other hand, the semantics encoded by the SAS and SASv2 methods are based on the YOLO detector. Consequently, the encoded information may not contain details such as interactions with the objects, but only which objects are present in the scene. For that reason, the distribution of the selected frames becomes more similar to a uniform selection since such objects tend to be present in most frames of the video. Besides, these encoders are limited to a small set of objects while natural language sentences present, potentially, a higher number of objects and interactions.

The BoT method did not perform well, although using the instructions as input. We argue that the BoT method depends on the accuracy of multiple external components such as the dense captioning and the video saliency algorithms; therefore, it is prone to errors. For that same reason, it yields the worst processing time among the methods. We noticed that the saliency is the major contributor for the BoT's semantic encoder. The errors on saliency estimation lead the method to present higher frame scores in segments that do not represent an instruction step (see images $2$ and $1$ in Figure~\ref{fig:qualitative_evaluation}-a and c, respectively) or disregard relevant segments. In many cases, higher saliency values are assigned to the person carrying out the task and lower at the task itself.

Besides outperforming the competitors, our method is also on par with FFNet, even presenting a better Recall in the YouCook2. Note that FFNet has no speed-up control optimization at the video level. Therefore, their agent can conveniently reduce and increase its frame skip according to the number of relevant frames present in the video. \Ie, output videos with more relevant frames present lower output speed-up rates while others with less relevant frames present higher rates. Moreover, in order to avoid abrupt speed-up rate changes along the video to keep transition smoothness, our agent is constrained via $\nu$ and $\omega$ to gradually change its navigation speed, while FFNet does not have this constraint.

To assess the significance of the results presented in Table~\ref{tab:main_results}, we ran a Student's $t$-test considering each metric and each pair of methods. We verified that, with $99\%$ confidence (\textit{p}-value~$<0.01$), our method is statistically superior to SAS, BoT, and SASv2 in Precision, Recall, and F1 metrics in both datasets. It also shows that the average values for F1 do not differ significantly when comparing our approach to FFNet. Our approach is statistically tied with the FFNet, and SAS methods concerning the OS metric in the YouCook2. We refer the reader to the supplementary material for tables presenting all the \textit{p}-values.

Figure~\ref{fig:barplot_ablation_speedup_progress} depicts our average OS error in the YouCook2 for different target rates (from ${2\times}$ to ${20\times}$), \ie, the $y$-axis shows ${\hat{S} - S^\ast}$. Note that the errors in the output rates are no higher than ${1.75\times}$, indicating that our agent can effectively control the output video's length without re-training. We accredit these results to our joint reward signal, which leads the agent to drop even the relevant frames in favor of balancing both objectives. The task of addressing the target speed-up becomes even more challenging when higher target values are imposed. This can also be observed in Figure~\ref{fig:barplot_ablation_speedup_progress}, where the agent creates output videos with less accuracy in their length as the target values increase.

\begin{figure*}[t!]
	\centering
	\includegraphics[width=\linewidth]{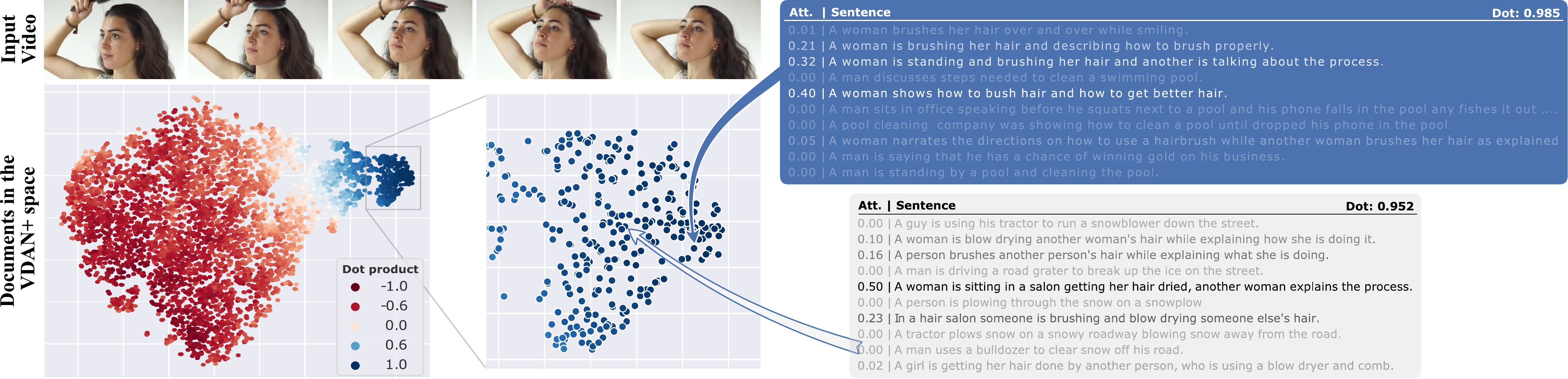}
	\caption{\textbf{Qualitative results for VDAN+}. The Figure shows VDAN+ document embeddings from the VaTeX validation set projected into two dimensions via t-SNE~\cite{vanDerMaaten2008} and colored according to their alignment with a given input video. Note that the documents with the highest alignment with the input video (blue points) are composed of semantically related sentences, and these sentences present the higher attention weights (Att.).}
	\label{fig:vdan_plus_tsne}
\end{figure*}
\subsubsection{Qualitative Results}

Figure~\ref{fig:qualitative_evaluation} shows qualitative results for three videos in the YouCook2 and COIN test sets. The vertical colored bars represent each method's selected frames, and the contiguous black blocks represent the ground-truth segments. Each method's $\hat{S}$ and ${S^\ast}$ are shown on the right. The numbered images outlined in green represent frames from segments where the agent takes the correct decision \wrt the ground-truth. Those outlined in red represent frames where the agent fails to increase its skip in segments with no recipe steps or decrease it in the ground-truth segments.

In general, SAFFA reduces its skip rate in the relevant segments and increases it otherwise. This behavior indicates that the VDAN+ space is useful for the agent to determine when the video segments match the input document. Some segments not in the ground-truth are semantically similar to the instructions, leading our agent to reduce its skip erroneously. That is the case illustrated in image $3$ in Figure~\ref{fig:qualitative_evaluation}-b. It is noteworthy that, to attend the target input speed-up rate, the agent decides to increase its skip rate even though the segment contains relevant frames. This is illustrated in images $4$ in Figure~\ref{fig:qualitative_evaluation}-a and c. However, note that image $4$ in Figure~\ref{fig:qualitative_evaluation}-c shows a transition scene where the person takes the tape to wind the junction. Although this scene is in the ground-truth segment, speeding it up does not incur losing the message conveyed by the video. Moreover, discarding relevant frames is necessary once a single ground-truth segment (\eg, third step in Figure~\ref{fig:qualitative_evaluation}-c) might represent a significant portion of the input video.

We assessed the usefulness of VDAN+ representation for the agent regarding the capability of navigating throughout videos. For each sample in the VaTeX validation set, we created a positive document, as described in Section~\ref{sec:methodology_training}, and paired it with a single video, \ie, the same video was used in all pairs. These pairs were projected into the VDAN+ space and further projected into a $2$-dimensional space via t-SNE~\cite{vanDerMaaten2008}. In Figure~\ref{fig:vdan_plus_tsne}-left, we illustrate the document embeddings (colored points) and some frames of the video. The colors represent the alignment between the documents and the video. On the right, we show the document that describes the video (shaded in blue) and a random document (shaded in gray) from a region that contains the documents most similar to the video. Note that the sentences in the documents with the highest attention weights are semantically related to the video. It indicates that our weakly-supervised approach is useful for the agent to navigate not only in instructional videos, as demonstrated in previous sections but also in videos from other domains.

\subsubsection{Ablation Studies}
\label{sec:ablation_studies}

\begin{figure}[t!]
	\centering
	\includegraphics[width=0.8\columnwidth]{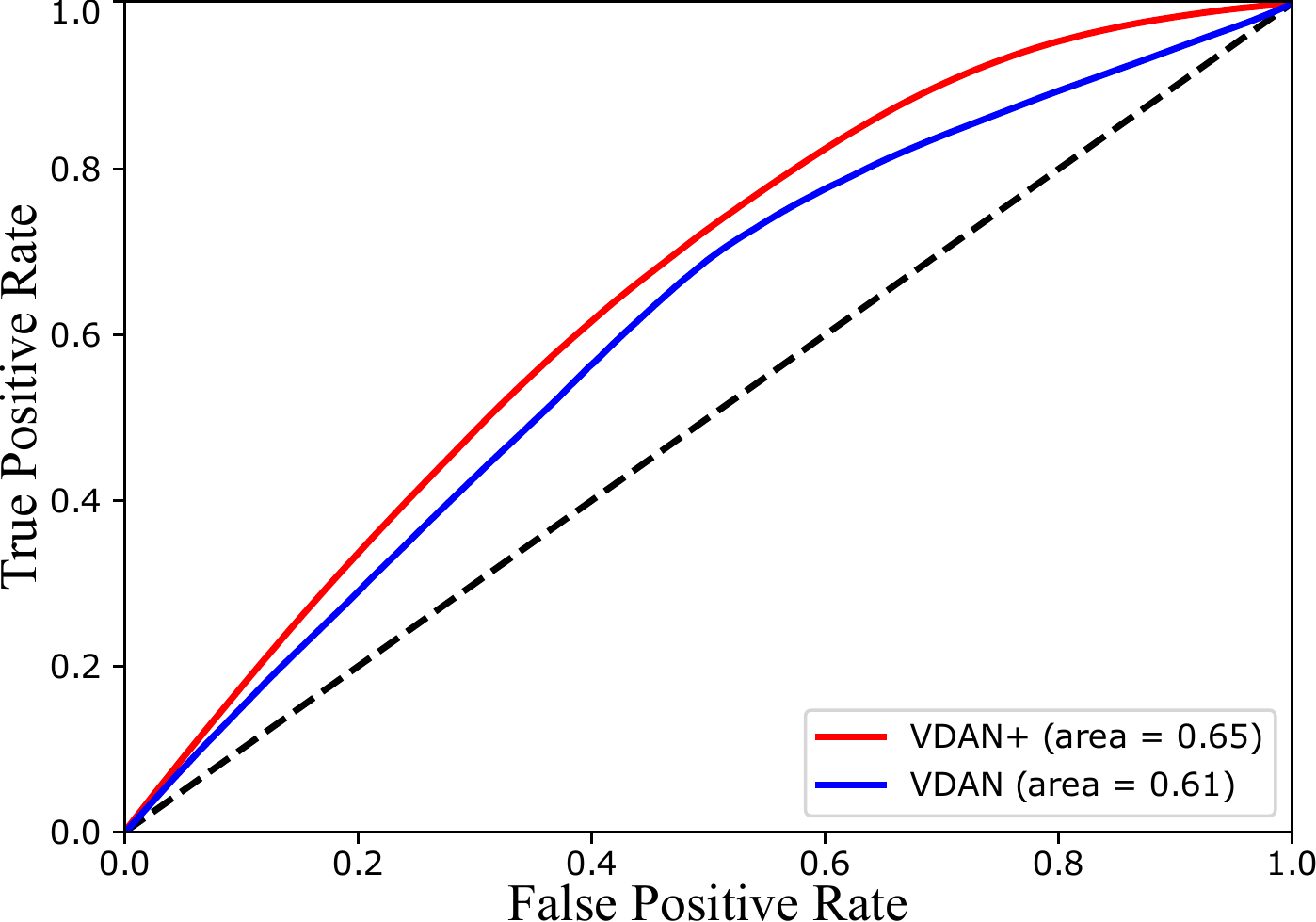}
	\caption{\textbf{VDAN+ versus VDAN~\cite{Ramos2020cvpr}.} The Receiver Operating Characteristic (ROC) curves in the YouCook2 when using all dot products as thresholds. VDAN+ is superior in deciding if a frame/segment is relevant.}
	\label{fig:vdan_vs_vdan_plus}
\end{figure}

\noindent \textbf{VDAN+ versus VDAN~\cite{Ramos2020cvpr}.} The embedding space created by the semantic encoder is crucial for the good performance of SAFFA since it directly observes the video and document features to infer the current semantic load. To demonstrate the superiority of the VDAN+ over VDAN~\cite{Ramos2020cvpr}, we approach the alignment of embeddings as a binary classification problem. First, we label the frames in the ground-truth segments of the YouCook2's training set as positive and the remaining ones as negative. Then, we apply multiple thresholds to the dot product values between each video frame/clip and the corresponding input recipe and calculate the Area Under Curve (AUC) computed from the Receiver Operating Characteristic (ROC) curve. For the experiments with the VDAN, we used the same hyperparameters and training data used in~\cite{Ramos2020cvpr}. Figure~\ref{fig:vdan_vs_vdan_plus} shows the ROC curve and the AUC scores for each semantic encoder. We see that VDAN+ is clearly superior to VDAN at correctly classifying a pair of frame/clip and document, yielding an AUC of $65$ against $61$ of VDAN. We claim that this superiority is related to the fact that VDAN+ creates embeddings at the segment level. Therefore, it avoids several false positives, such as correlating static scenes and objects under no action.

\noindent \textbf{The effect of ${\mathbf{h}_0 = \phi(v)}$.} Apart from initializing the first hidden state vector of our semantic encoder as ${\mathbf{h}_0 = \phi(v)}$, we evaluated two other variants: ${\mathbf{h}_0 = \mathbf{0}}$ and; ${\mathbf{h}_0 = W_s}$, where ${W_s \in \mathbb{R}^z}$ is a learnable parameter vector. We computed the Mean Reciprocal Rank (MRR) using the VaTeX validation set as described in Section~\ref{sec:methodology_training} to compare the variants with our approach.

The MRR is given by $\frac{1}{Q}\sum_{q=1}^{Q}\frac{1}{rank_q}$, where $Q$ is the number of documents composing the query set and $rank_q$ is the rank at which the relevant document was retrieved. As expected, the MRR values for ${\mathbf{h}_0 = \phi(v)}$ are far superior when compared to both ${\mathbf{h}_0 = \mathbf{0}}$ and ${\mathbf{h}_0 = W_s}$ variants, presenting $0.718$, $0.065$, and $0.061$, respectively. We observed that using any of the two variants degraded the model's capacity to produce semantically relevant $\mathbf{e}_D$ vectors. Basically, after training for $100$ epochs, the attention module could still not attend the correct sentences because the attention weights became roughly evenly distributed. This outcome was expected since $\phi(v)$ lies in a well-defined semantic space, making it easier to create $\mathbf{d}$ given the trajectory $\{{\mathbf{h}_0=\phi(v)}, \mathbf{h}_1, \dots, \mathbf{h}_N\}$ in the $512$-dimensional space.

\noindent \textbf{Method's composition.} We evaluated the impact of using different configurations to integrate a variant of our approach. Table~\ref{tab:ablation_agent} presents the results for Precision, Recall, F1, OS, and OP. The column SE shows the Semantic Encoder used; SA shows the Skip-Aware component, which indicates the agent's current average skip rate; and the NRPE is the Normalized Reversed Positional Encoding, which informs the agent of its relative location in the video. All agent variations were trained with the same hyperparameters as detailed in Section~\ref{sec:implementation_details}, except for the terminal reward weight, $\lambda$ (Equation~\ref{eq:immediate_reward}). In the variations using VDAN as the semantic encoder, we set ${\lambda=F}$ since we observed that local rewards were dominating the signal sent to the agent. It is noteworthy that the only difference of the SAFFA-based variants without SA and NRPE to the method proposed in our previous work~\cite{Ramos2020cvpr} is in the reward. The SFF-RL agent disregards the speed-up control, \ie, ${\lambda=0}$. We also included a variant that uses the BoT's frame sampler, a non-RL solution. We used the frames and document alignments throughout the video as input.

As can be seen in Table~\ref{tab:ablation_agent}, the SA element is crucial for the agent to attend the target speed-up rate; note the low OP values. In the cases without SA and NRPE, due to the lack of distinction among states with the same semantic features, the agent ends up taking the same actions under varied conditions, \eg, at the beginning/end of the video or with the average speed-up rate close/far from the ideal. Adding only the NRPE is insufficient since the agent remains ``blind'' about its skip rate. It is worth noting that using different encoders produced oppositive behaviors \wrt the OS metric. These results relate to the distribution of VDAN and VDAN+ embeddings, which present average dot products of ${0.175\pm0.812}$ and ${0.833\pm0.358}$ for VDAN and VDAN+, respectively, in the YouCook2 training set. The higher semantic rewards of VDAN+ lead the agent to reduce its skip rate and accumulate more rewards.

The variations with SA vector and no NRPE aid the agent in attending the target speed-up rate. However, in general, the agent becomes cautious in terms of its skip rate, preferring not to intensively increase or decrease the video's speed-up rate to favor the relevance of the frames. For comparison, the agents' F1 Score using this state composition is $14.09$ with VDAN and $15.18$ with VDAN+, while $14.33$ and $17.86$ when using all the state's components. The agent using all components proposed in this paper presents the best trade-off between relevance and target speed-up rate in the output videos as confirmed by the OP metric. By using the information of its position in the video, the agent can freely decide to attend to the semantics objective at the beginning of the video and increase its concern about the speed-up rate as it reaches the end of the video.

It is worth mentioning the Non-RL variant result. Its F1, OS, and OP values are better than those achieved by the VDAN-based variants, confirming the superiority of VDAN+ over VDAN. Although this variant is on par with the ``VDAN+ with SA'', the BoT frame sampler takes over $60\times$ more processing time.

\begin{table}[t!]
	\centering
	\caption{\textbf{Ablation Study.} Precision, Recall, F1 Score, Output Speed-up (OS), and Overall Performance (OP) in YouCook2, \wrt the method's compo- sition. SE stands for Semantic Encoder, SA for Skip-Aware, NRPE for Normalized Reversed Positional Encoding, and FS for Frame Sampler.}
	\label{tab:ablation_agent}
	\setlength{\tabcolsep}{1.5pt}
	\scriptsize{
	\begin{tabular}{lccccccccccc}
		\toprule
		\multicolumn{4}{c}{\thead{\scriptsize{\textbf{Method's Composition}}}} & & \multicolumn{5}{c}{\thead{\scriptsize{\textbf{Metrics}}}} \\ 
		\multicolumn{1}{c}{SE} & SA & NRPE & \multicolumn{1}{c}{FS} & & Precision$^1$ &  Recall$^1$ &  F1 Score$^1$ & OS$^2$ & OP$^1$	\\  
		\cmidrule{1-4} \cmidrule{6-10}
		VDAN & \xmark & \xmark & SAFFA & & $ 49.17 $ & $ 04.00 $ & $ 07.34 $ & $ 24.87 $ & $ 00.00 $ 	\\
		VDAN & \xmark & \cmark & SAFFA & & $ 49.68 $ & $ 04.26 $ & $ 07.77 $ & $ 23.67 $ & $ 00.00 $ 	\\
		VDAN & \cmark & \xmark & SAFFA & & $ 49.32 $ & $ 08.37 $ & $ 14.09 $ & $ 11.93 $ & $ 24.69 $ 	\\
		VDAN & \cmark & \cmark & SAFFA & & $ 51.84 $ & $ 08.45 $ & $ 14.33 $ & $ 12.42 $ & $ 24.79 $ 	\\
		\cdashlinelr{1-10}
		VDAN+ & \xmark & \xmark & SAFFA & & $ 59.11 $ & $ 79.08 $ & $ 64.96 $ & $ 01.94 $ & $ 00.00 $	\\
		VDAN+ & \xmark & \cmark & SAFFA & & $ 59.63 $ & $ 47.67 $ & $ 50.98 $ & $ 03.02 $ & $ 00.00 $ 	\\
		VDAN+ & \cmark & \xmark & SAFFA & & $ 52.39 $ & $ 09.03 $ & $ 15.18 $ & $ 11.86 $ & $ 26.33 $ 	\\
		VDAN+ & \cmark & \cmark & SAFFA & & $ 53.20 $ & $ 12.81 $ & $ 17.86 $ & $ 11.68 $ & $ 30.07 $ 	\\
		\cdashlinelr{1-10}
		VDAN  & \xmark & \xmark & SFF-RL~\cite{Ramos2020cvpr} & & $ 53.78 $ & $ 42.07 $ & $ 41.55 $ & $ 06.09 $ & $ 00.00 $ 	\\
		VDAN+ & \textcolor{gray}{N/A} & \textcolor{gray}{N/A}	 & Non-RL~\cite{Ramos2020wacv} & & $ 52.00 $ & $ 09.33 $ & $ 15.56 $ & $ 11.48 $ & $ 26.42 $ 	\\		
		\cmidrule{6-10}
        & & & & & \multicolumn{5}{r}{\scriptsize{\textit{$^1$Higher is better (\%)} \textit{$^2$Better closer to ${S^\ast=12}$}}} \\
		\bottomrule 
	\end{tabular}
	}
\end{table}

\begin{figure}[t!]
	\centering
	\includegraphics[width=\linewidth]{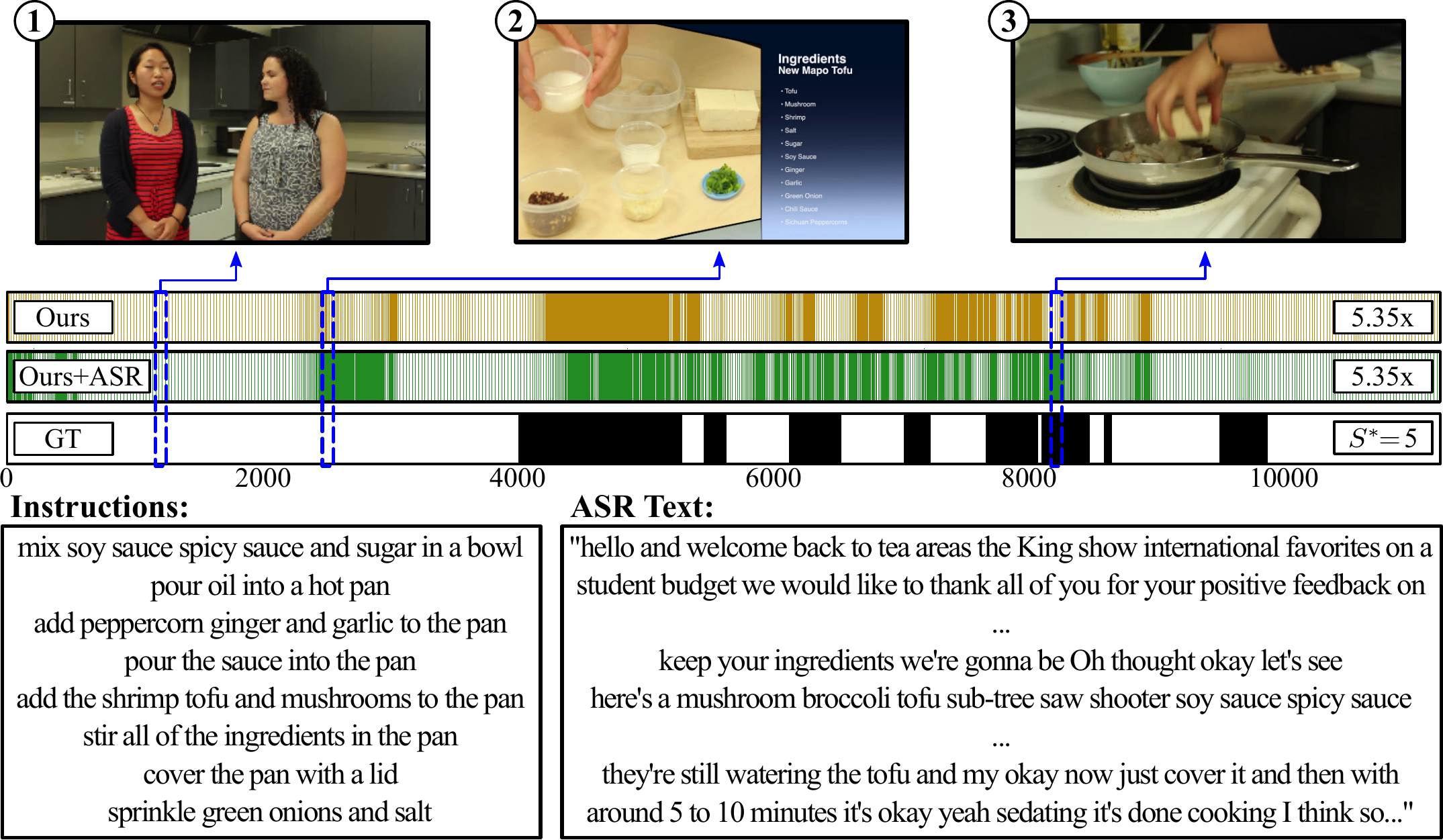}
	\caption{\textbf{Robustness to Automatic Speech Recognition (ASR).} The image illustrates the difference between using the annotated recipe and the ASR text as input. Note that the distribution of the selected frames is similar for either input (colored bars represent the selected frames).}
	\label{fig:ablation_asr}
\end{figure}

\noindent \textbf{Robustness to Automatic Speech Recognition (ASR).} A representative aspect of instructional videos is that the person recording the instructions usually narrates the steps being performed. Thus, using an ASR system, one could utilize our methodology to create an accelerated video with no need for text input by the user. To demonstrate this application, we selected a video from the YouCook2's validation set and automatically created a document using the English subtitles generated by the YouTube ASR system. We ran our method using that input and compared it against our main result. Figure~\ref{fig:ablation_asr} shows the result (${S^\ast=5}$) with some representative frames on top, the frame selection in the middle, and the input documents for each approach at the bottom.

In general, our method performed a similar frame sampling using either input document. The reason is that, although the input documents are different in length, they have similar semantics in content. In Figure~\ref{fig:ablation_asr}, images 1 and 3 illustrate some cases where our method was accurate \wrt the ground-truth. It keeps a sparser frame selection in the recorder's introductory speech (see image $1$) and a denser frame selection when the food is cooked (see image $3$). We see a difference between the frame selection when using the instructions and the recorder speech in the region illustrated in image $2$. SAFFA using ASR decided to reduce the skip rate in the segment that does not represent a recipe step, but notice that it presents the ingredients used. This behavior is comprehensible since the ingredients are described in some sentences.

\section{Discussions}
\label{sec:discussions}

In this paper, we proposed a novel methodology based on a reinforcement learning formulation to accelerate instructional videos. The agent, namely Skip-Aware Fast-Forwarding Agent (SAFFA), is trained to decide which frames to remove based on textual data and a target speed-up rate. We also presented an extension to the Visually-guided Document Attention Network (VDAN)~\cite{Ramos2020cvpr}, called VDAN+, which creates a highly discriminative embedding space to represent both textual and visual data. Our approach outperforms several competitors in F1 Score while allowing the user to define the video output length, which can be crucial for applications where time and storage resources are scarce.

\begin{figure}[t!]
	\centering
	\includegraphics[width=\linewidth]{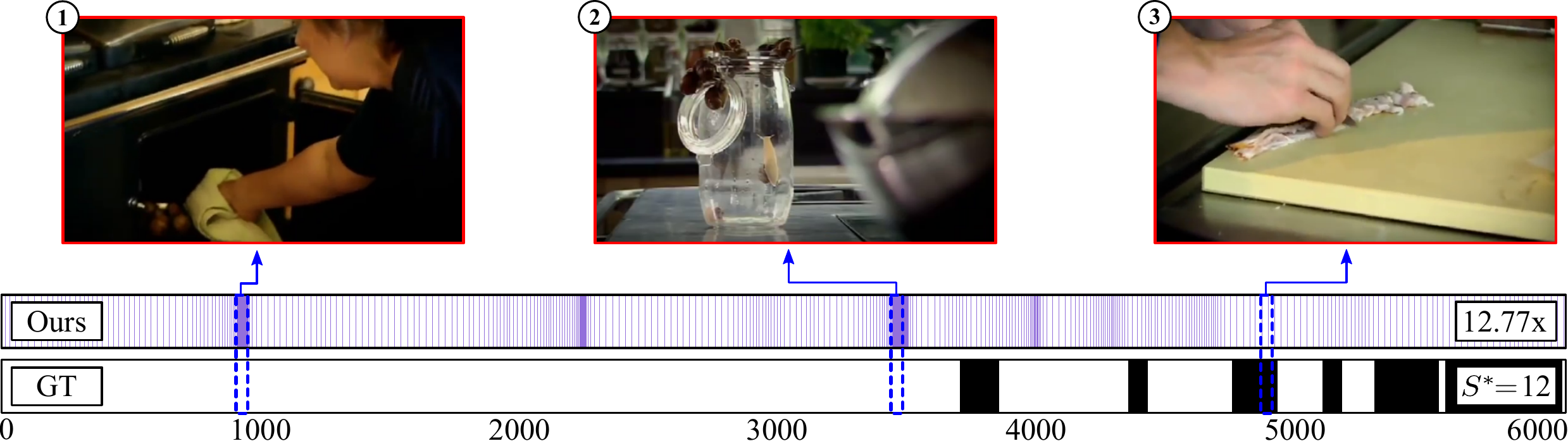}
	\caption{\textbf{Limitations.} The colored bars, contiguous black blocks, and images outlined in red represent, respectively, the selected frames, the ground-truth, and the regions where the agent took a wrong decision.}
	\label{fig:limitations}
\end{figure}

Despite achieving the best results, our methodology suffers from certain limitations. The agent may take incorrect actions and emphasize segments that are not the recipe's steps. Figure~\ref{fig:limitations} illustrates an example of such a case, where the agent reduces its skip rate to emphasize a person putting food in the oven (image 1) and some snails getting out of the recipient (image 2). The major reason for this outcome is the design of our reward function that is sparse \wrt the speed-up rate deviation. Thus, at the beginning of the video, the sum of discounted rewards will mostly be composed of semantic signals, while as $t$ gets higher, the speed-up deviation signal gets stronger. Suppose the agent mistakenly reduces its skip rate at the beginning of the video due to a potential relevant segment. In that case, it must skip the final frames even if they are also relevant (image 3). In this example, VDAN+ contributed negatively, producing highly aligned vectors due to the high semantic relation between the segments represented in images 1 and 2 with instructions like ``Remove the snail from the shell'' and ``Chop and cook the pancetta in a pan'' of the recipe.

Nonetheless, our approach takes a step towards understanding the training of RL agents in a multi-modal environment. To accelerate long untrimmed videos emphasizing the relevant content, we trained an agent to collect rewards proportional to video-text similarity powered by VDAN+, which removed the demand of temporal markings. This strategy allows the agent to run with freely available texts on the web and requires fewer annotation efforts. We believe that the results may benefit further works in the direction of weakly supervised reinforcement learning in the case of availability of multi-modal data and scarcity of annotations.

\ifCLASSOPTIONcompsoc
  \section*{Acknowledgments}
\else
  \section*{Acknowledgment}
\fi

We thank the agencies CAPES, CNPq, FAPE\-MIG, and Petrobras for funding different parts of this work. We also thank NVIDIA Corporation for the donation of a GPU.

\ifCLASSOPTIONcaptionsoff
  \newpage
\fi



\bibliographystyle{IEEEtran}
\bibliography{IEEEabrv,Main}
\vspace{-1cm}
\begin{IEEEbiography}[{\includegraphics[width=1in,height=1.25in,clip,keepaspectratio]{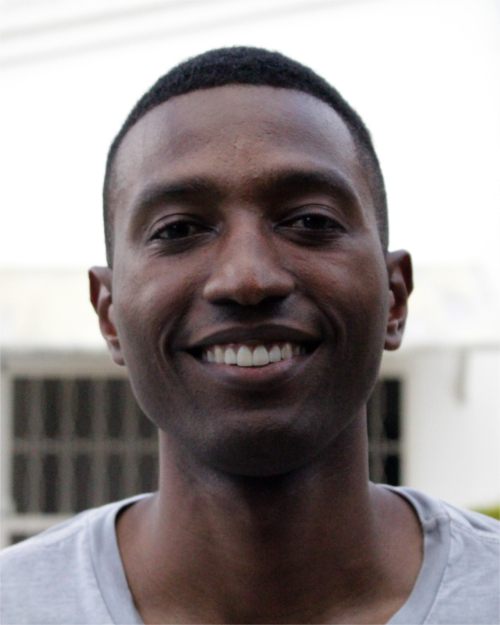}}]{Washington Ramos}
received M.S. degree in Computer Science from the Universidade Federal de Minas Gerais (UFMG), Brazil, and B.S. degree in Computer Science from the Pontifícia Universidade Católica de Minas Gerais (PUC-Minas), Brazil. He is currently a Ph.D. student in Computer Science in the Vision and Robotics Laboratory (VeRLab) at UFMG. His research interests are egocentric vision, vision-language modeling, and reinforcement learning.%
\end{IEEEbiography}
\vspace{-1cm}
\begin{IEEEbiography}[{\includegraphics[width=1in,height=1.25in,clip,keepaspectratio]{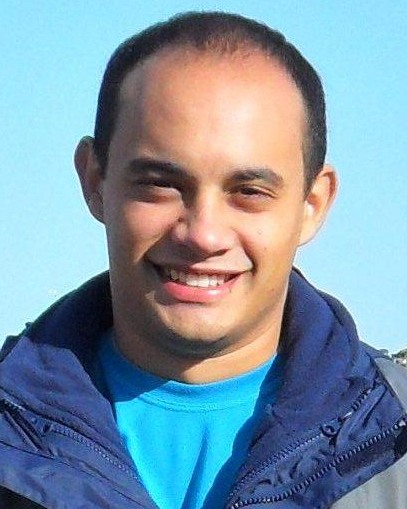}}]{Michel Silva}
is an Assistant Professor at Universidade de Viçosa (UFV) and researcher at the Vision and Robotics Laboratory (VeRLab), Universidade Federal de Minas Gerais (UFMG), Brazil. He received Ph.D. degree in Computer Science from UFMG, and B.S. and M.S. degrees from Universidade Federal de Lavras (UFLA), with a visiting student period at James Hutton Institute (JHI), UK. His interest is on computer vision, with focus on egocentric videos and applications.
\end{IEEEbiography}
\vspace{-1cm}
\begin{IEEEbiography}[{\includegraphics[width=1in,height=1.25in,clip,keepaspectratio]{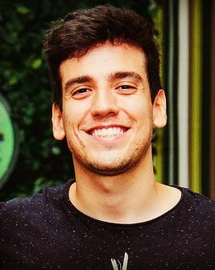}}]{Edson Araujo} received his B.S. degree in Computer Science from the Universidade Federal de Minas Gerais (UFMG), Brazil. He is currently a M.S. student in Computer Science in the Vision and Robotics Laboratory (VeRLab) at UFMG. His research interests are egocentric vision, multimodal learning, self-supervised learning, and medical image analysis.
\end{IEEEbiography}
\vspace{-1cm}
\begin{IEEEbiography}[{\includegraphics[width=1in,height=1.25in,clip,keepaspectratio]{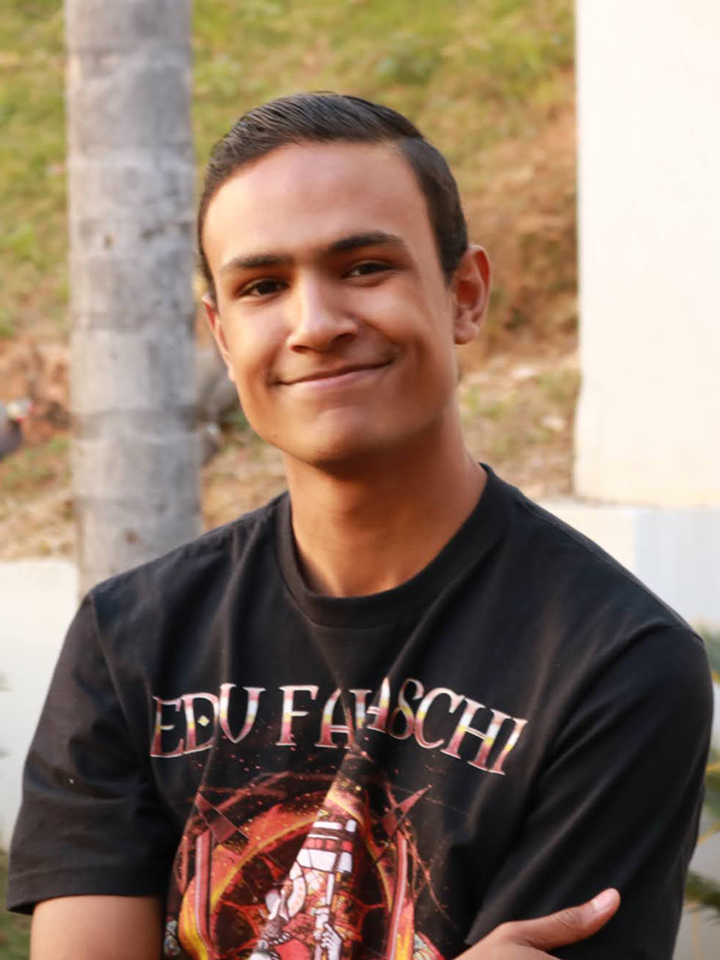}}]{Victor Moura} is a B.S. student in Computer Science at the Universidade Federal de Minas Gerais (UFMG), Brazil, and undergraduate researcher in the Vision and Robotics Laboratory (VeRLab) at UFMG. His research interests include computer graphics, computer vision, and three-dimensional reconstruction.
\end{IEEEbiography}
\vspace{-1cm}
\begin{IEEEbiography}[{\includegraphics[width=1in,height=1.25in,clip,keepaspectratio]{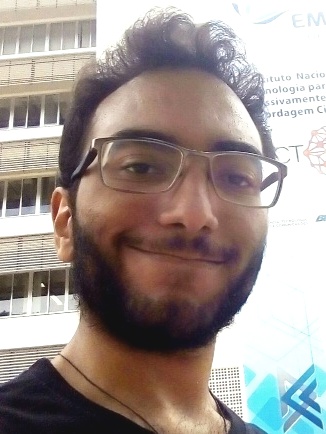}}]{Keller Oliveira} is a B.S. student in Computer Science at the Universidade Federal de Minas Gerais (UFMG), Brazil, and undergraduate researcher in the Vision and Robotics Laboratory (VeRLab) at UFMG. His research interests include computer graphics, computer vision and machine learning applied to entertainment media.
\end{IEEEbiography}
\vspace{-1cm}
\begin{IEEEbiography}[{\includegraphics[width=1in,height=1.25in,clip,keepaspectratio]{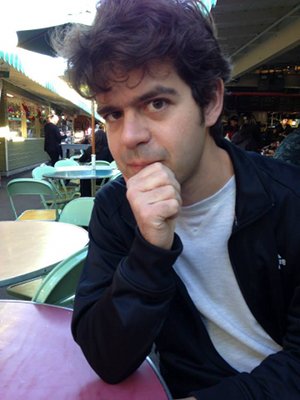}}]{Leandro Soriano Marcolino} is a lecturer (assistant professor) at Lancaster University. He obtained his PhD at University of Southern California (USC). He received the best dissertation and the best research assistant award from the Computer Science Department at USC, had a paper nominated for best paper from the leading multi-agent conference AAMAS, and had his undergraduate work selected as the best in the nation by the Brazilian Computer Science Society. His research interests are multi-agent teamwork, machine learning, games, and robotics; with emphasis on coordination and collaboration.
\end{IEEEbiography}
\vspace{-1cm}
\begin{IEEEbiography}[{\includegraphics[width=1in,height=1.25in,clip,keepaspectratio]{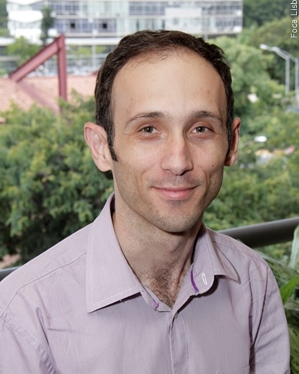}}]{Erickson R. Nascimento} is an Associate Professor in the Department of Computer Science at the Universidade Federal de Minas Gerais (UFMG), Brazil. He holds B.S., M.S. and Ph.D degrees in Computer Science from UFMG. He was also a visiting researcher at Berkeley Artificial Intelligence (BAIR) lab of the UC Berkeley. 
His research interests include Computer Vision and Pattern Recognition. In particular,  egocentric videos, multimodal learning, and local feature extraction.%
\end{IEEEbiography}

\newpage


\section*{Supplementary material}

# [Supplementary Material]
# Text-Driven Video Acceleration: A Weakly-Supervised Reinforcement Learning Method

Washington Ramos [ID] Michel Silva [ID] Edson Araujo [ID], Victor Moura [ID], Keller Oliveira [ID], Leandro Soriano Marcolino [ID], and Erickson R. Nascimento [ID]

---

I N this supplementary material, we present details about the baselines, extra qualitative results with discussion, and the complete results for the statistical significance test for Precision, Recall, F1 Score, and Output Speed-up yielded by each method. In addition, we present details for the processing time of the methods.

## 1 BASELINES DETAILS

We prepared the input data for each baseline method as follows:

- **SAS [1] and SASv2 [2].** In the YouCook2 dataset, we adapted the SAS and SASv2 methods to use food-related content as the semantic input. In particular, we filtered their content descriptor based on the YOLO classes to use only kitchen-related objects (*e.g.*, cup, orange, carrot, fork, *etc.*). In the COIN dataset, we used a different filtering strategy. For each domain, we ranked the YOLO objects by their GloVe embeddings similarity with the domain name and selected the top 8 (10%) as the semantic input. For domain names with compound names, we used the average embedding. Table 1 show the final set of objects for each domain. These filterings ensures that the methods would work better for each class of videos and make these baselines more robust.
- **FFNet [3].** For the FFNet, we annotate all frames in the YouCook2 and COIN training sets as relevant or not based on the temporal boundaries of the instruction steps. These annotations are only used to feed the FFNet agent during its training; they are not used in SAFFA and its variants either during training or testing.

---

- *Washington Ramos, Edson Araujo, Victor Moura, Keller Oliveira and Erickson R. Nascimento are with the Computer Science Deparment, Universidade Federal de Minas Gerais, Belo Horizonte, MG 31270-901, Brazil. E-mail: {washington.ramos, edsonroteia, victorhugomoura, kellermartins, erickson}@dcc.ufmg.br*
- *Michel Silva is with the Department of Informatics, Universidade Federal de Viçosa, Viçosa, MG 36570-900, Brazil. E-mail: michel.m.silva@ufv.br*
- *Leandro Soriano Marcolino is with Lancaster University, Lancaster, LA1 4YW, U.K. E-mail: l.marcolino@lancaster.ac.uk*

- **Other approaches.** We concatenated all the instruction steps into a single document to be used as input to the BoT [4] and our methodology.

## 2 STATISTICAL TESTS

Tables 2 and 3 show the $p$-values obtained in the Student's $t$-test, which were summarized in the main text, given each pair of methods. For the test, we used a significance of $\alpha = 0.01$, which indicates that $p$-values above $0.01$ are strong evidence that a pair of methods is unlikely to produce statistically different results, given a metric being assessed. We highlight the $p$-values above $0.01$ with a bold font.

## 3 QUALITATIVE RESULTS

Figure 1 presents extra qualitative results regarding the main text composed of 2 videos in each test set of YouCook2 and COIN datasets. The vertical colored bars inside the black bounding boxes represent the selected frames of each method. The contiguous black blocks represent the ground-truth segments. Each method's output speed-up rate for a given video is shown on the right of the corresponding bounding box. The target speed-up of each video is represented inside the GT bounding box. The numbered images outlined in green represent frames from segments where the agent takes the correct decision w.r.t. the ground-truth. Those outlined in red represent frames where the agent fails to increase its skip in segments with no instruction steps or decrease it in the ground-truth segments.

According to the results presented in the main text, when analyzing the ones presented in this document, we can affirm that our agent reduces its skip rate in the relevant segments and increases it otherwise, in general. Moreover, the agent sped up the moments where the presenters are in focus, as presented in images 3 and 2 of Figure 1-a and b, respectively. Usually, those moments are not strictly related to instructional steps. However, this behavior led our agent to accelerated erroneously the clip presented by image 2 of Figure 1-c, which contains the presenter in focus and

TABLE 1
Most related YOLO objects per COIN domain.

| COIN Domain | Associated YOLO objects |
|---|---|
| Nursing and Care | ['person', 'bed', 'chair', 'stopsign', 'umbrella', 'clock', 'bear', 'cat'] |
| Vehicles | ['car', 'truck', 'bus', 'motorbike', 'parking', 'bicycle', 'train', 'person'] |
| Leisure and Performance | ['person', 'book', 'table', 'tie', 'traffic', 'sink', 'parking', 'umbrella'] |
| Gadgets | ['laptop', 'toaster', 'backpack', 'refrigerator', 'phone', 'sink', 'scissors', 'keyboard'] |
| Electrical Appliances | ['refrigerator', 'toaster', 'toothbrush', 'laptop', 'sink', 'microwave', 'phone', 'handbag'] |
| Furniture and Decoration | ['glass', 'table', 'chair', 'sofa', 'sink', 'handbag', 'vase', 'person'] |
| Science and Craft | ['book', 'chair', 'person', 'sink', 'boat', 'stopsign', 'umbrella', 'train'] |
| Plants and Fruits | ['plant', 'banana', 'bear', 'carrot', 'sink', 'orange', 'refrigerator', 'apple'] |
| Drink and Snacks | ['pizza', 'bottle', 'cake', 'sandwich', 'refrigerator', 'person', 'banana', 'table'] |
| Dish | ['cake', 'spoon', 'pizza', 'microwave', 'sandwich', 'refrigerator', 'oven', 'table'] |
| Sports | ['baseball bat', 'baseball glove', 'car', 'person', 'bicycle', 'horse', 'motorbike', 'book'] |
| Housework | ['refrigerator', 'sink', 'hotdog', 'handbag', 'toothbrush', 'skateboard', 'scissors', 'giraffe'] |

TABLE 2
**$p$-values obtained in the Student's $t$-test in the YouCook2 dataset**.
Values above $\alpha = 1.00e{-}02$ (in bold) indicate that a pair of methods is unlikely to produce statistically different results, given the metric.

**PRECISION**

| | FFNet [3] | SAS [1] | SASv2 [2] | BoT [4] | Ours |
|---|---|---|---|---|---|
| FFNet [3] | **1.00e+00** | 4.26e−17 | 3.47e−16 | 1.18e−16 | 5.42e−07 |
| SAS [1] | 4.26e−17 | **1.00e+00** | **6.69e−01** | **4.97e−01** | 4.21e−04 |
| SASv2 [2] | 3.47e−16 | **6.69e−01** | **1.00e+00** | **2.83e−01** | 1.52e−03 |
| BoT [4] | 1.18e−16 | **4.97e−01** | **2.83e−01** | **1.00e+00** | 1.06e−04 |
| Ours | 5.42e−07 | 4.21e−04 | 1.52e−03 | 1.06e−04 | **1.00e+00** |

**RECALL**

| | FFNet [3] | SAS [1] | SASv2 [2] | BoT [4] | Ours |
|---|---|---|---|---|---|
| FFNet [3] | **1.00e+00** | 2.45e−28 | 2.39e−11 | 3.49e−27 | **1.72e−01** |
| SAS [1] | 2.45e−28 | **1.00e+00** | 3.49e−37 | **1.06e−01** | 2.11e−07 |
| SASv2 [2] | 2.39e−11 | 3.49e−37 | **1.00e+00** | 1.04e−19 | 2.56e−04 |
| BoT [4] | 3.49e−27 | **1.06e−01** | 1.04e−19 | **1.00e+00** | 5.55e−08 |
| Ours | **1.72e−01** | 2.11e−07 | 2.56e−04 | 5.55e−08 | **1.00e+00** |

**F1 SCORE**

| | FFNet [3] | SAS [1] | SASv2 [2] | BoT [4] | Ours |
|---|---|---|---|---|---|
| FFNet [3] | **1.00e+00** | 1.03e−31 | 3.94e−13 | 1.09e−29 | **1.30e−01** |
| SAS [1] | 1.03e−31 | **1.00e+00** | 9.28e−28 | 8.37e−02 | 1.41e−08 |
| SASv2 [2] | 3.94e−13 | 9.28e−28 | **1.00e+00** | 5.81e−17 | 5.36e−03 |
| BoT [4] | 1.09e−29 | 8.37e−02 | 5.81e−17 | **1.00e+00** | 1.30e−09 |
| Ours | **1.30e−01** | 1.41e−08 | 5.36e−03 | 1.30e−09 | **1.00e+00** |

**OUTPUT SPEED-UP**

| | FFNet [3] | SAS [1] | SASv2 [2] | BoT [4] | Ours |
|---|---|---|---|---|---|
| FFNet [3] | **1.00e+00** | **1.74e−01** | 2.51e−15 | **1.13e−01** | **3.57e−01** |
| SAS [1] | **1.74e−01** | **1.00e+00** | 0.00e+00 | 1.44e−04 | **7.79e−01** |
| SASv2 [2] | 2.51e−15 | 0.00e+00 | **1.00e+00** | 5.50e−27 | 1.52e−03 |
| BoT [4] | **1.13e−01** | 1.44e−04 | 5.50e−27 | **1.00e+00** | 6.27e−03 |
| Ours | **3.57e−01** | **7.79e−01** | 1.52e−03 | 6.27e−03 | **1.00e+00** |

TABLE 3
**$p$-values obtained in the Student's $t$-test in the COIN dataset**.
Values above $\alpha = 1.00e{-}02$ (in bold) indicate that a pair of methods is unlikely to produce statistically different results, given the metric.

**PRECISION**

| | FFNet [3] | SAS [1] | SASv2 [2] | BoT [4] | Ours |
|---|---|---|---|---|---|
| FFNet [3] | **1.00e+00** | 3.31e−10 | 3.94e−09 | 1.32e−10 | 1.42e−03 |
| SAS [1] | 3.31e−10 | **1.00e+00** | **6.26e−01** | **6.56e−01** | 1.90e−03 |
| SASv2 [2] | 3.94e−09 | **6.26e−01** | **1.00e+00** | **3.60e−01** | 7.78e−03 |
| BoT [4] | 1.32e−10 | **6.56e−01** | **3.60e−01** | **1.00e+00** | 6.69e−04 |
| Ours | 1.42e−03 | 1.90e−03 | 7.78e−03 | 6.69e−04 | **1.00e+00** |

**RECALL**

| | FFNet [3] | SAS [1] | SASv2 [2] | BoT [4] | Ours |
|---|---|---|---|---|---|
| FFNet [3] | **1.00e+00** | 9.71e−32 | 4.12e−24 | 1.45e−28 | 3.55e−03 |
| SAS [1] | 9.71e−32 | **1.00e+00** | 2.18e−03 | **6.00e−01** | 3.62e−37 |
| SASv2 [2] | 4.12e−24 | 2.18e−03 | **1.00e+00** | **2.43e−02** | 1.89e−30 |
| BoT [4] | 1.45e−28 | **6.00e−01** | **2.43e−02** | **1.00e+00** | 1.25e−34 |
| Ours | 3.55e−03 | 3.62e−37 | 1.89e−30 | 1.25e−34 | **1.00e+00** |

**F1 SCORE**

| | FFNet [3] | SAS [1] | SASv2 [2] | BoT [4] | Ours |
|---|---|---|---|---|---|
| FFNet [3] | **1.00e+00** | 1.65e−24 | 1.83e−18 | 2.79e−25 | **3.69e−01** |
| SAS [1] | 1.65e−24 | **1.00e+00** | **1.27e−02** | **5.40e−01** | 3.93e−21 |
| SASv2 [2] | 1.83e−18 | **1.27e−02** | **1.00e+00** | 3.22e−03 | 2.35e−15 |
| BoT [4] | 2.79e−25 | **5.40e−01** | 3.22e−03 | **1.00e+00** | 6.78e−22 |
| Ours | **3.69e−01** | 3.93e−21 | 2.35e−15 | 6.78e−22 | **1.00e+00** |

**OUTPUT SPEED-UP**

| | FFNet [3] | SAS [1] | SASv2 [2] | BoT [4] | Ours |
|---|---|---|---|---|---|
| FFNet [3] | **1.00e+00** | **1.82e−01** | 7.32e−22 | **1.50e−01** | 1.53e−07 |
| SAS [1] | **1.82e−01** | **1.00e+00** | 9.96e−18 | **9.30e−01** | 3.82e−05 |
| SASv2 [2] | 7.32e−22 | 9.96e−18 | **1.00e+00** | 4.80e−18 | 2.52e−04 |
| BoT [4] | **1.50e−01** | **9.30e−01** | 4.80e−18 | **1.00e+00** | 4.10e−05 |
| Ours | 1.53e−07 | 3.82e−05 | 2.52e−04 | 4.10e−05 | **1.00e+00** |

belongs to the ground-truth segments. In the case illustrated in image 4 of Figure 1-b, although the video segment is not in the ground-truth, it is semantically similar to the input instructions, which leads the agent to reduce the video speed. As mentioned in the main text, in some cases, the ground-truth might represent a significant portion of the input video, *e.g.*, the third step of Figure 1-a. Image 4 of this Figure depicts the agent deciding to increase the skip rate even though the segment contains relevant frames in order to attend the target input speed-up rate.

## 4 TIME ANALYSIS

In the main text, we presented in Table 1 the estimated processing times for each method. For estimating the values, we used an Intel® Core™ i7-3770K CPU @ 3.50GHz machine with 32GB RAM and an NVIDIA Titan RTX GPU to run all competitors in a set of selected videos from the YouCook2

**a) No Oil Hummus (YouCook2)**

**Instructions:**

1- pour drained liquid from tinned beans and chickpeas to a bowl
2- put the beans and chickpeas in a food processor
3- add 2 cloves of garlic lemon tahini salt and pepper to the food processor
4- add the liquid to the food processor
5- blend everything in the food processor

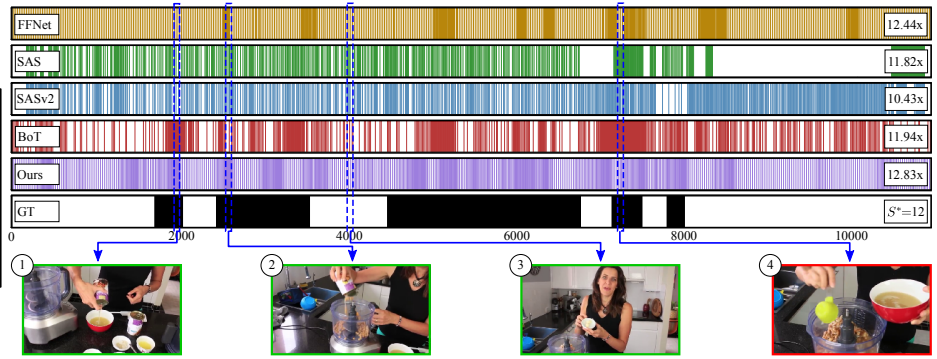

**b) Thai Fried Rice with Shrimp (YouCook2)**

**Instructions:**

1- add fish sauce to a bowl
2- slice chilis and a lime and add the pieces and juice to the bowl
3- peel the shrimp
4- chop the garlic onion green onion and chinese broccoli
5- heat up oil in a wok
6- fry garlic and shrimp in the wok
7- mix rice and an egg with the shrimp
8- add soy sauce oyster sauce and sugar to the wok
9- add the onion and broccoli to the wok
10- add the green onions to the wok

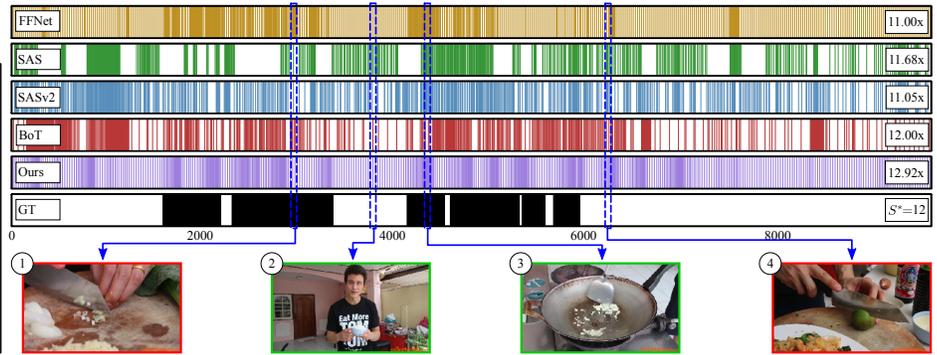

**c) Sow (COIN)**

**Instructions:**

1- dig some holes on the soil
2- sow on the soil
3- cover with some soil
4- water and fertilize the seeds
5- apply a cover on the soil

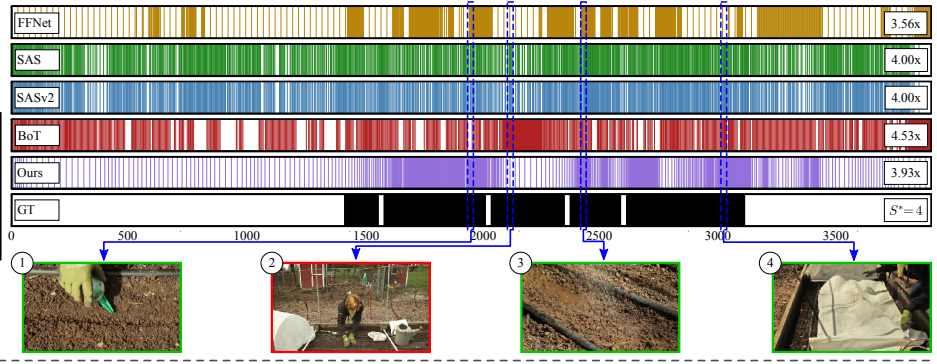

**d) How to Transplant a Tree (COIN)**

**Instructions:**

1- take out the plant
2- take out the plant
3- put in the plant
4- fill with some soil
5- fill with some soil

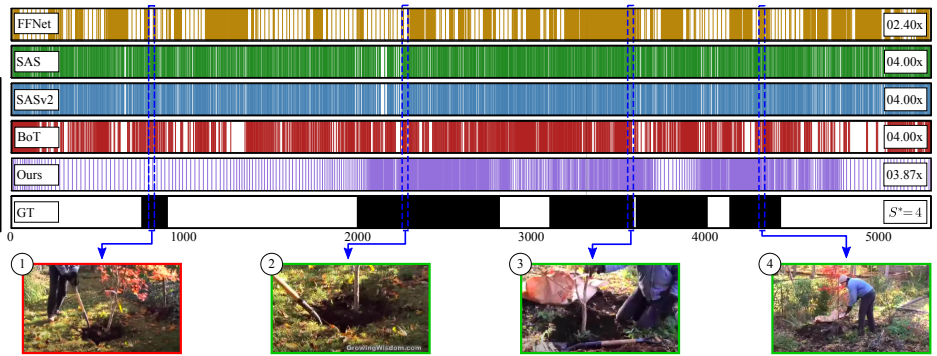

Fig. 1. **Extra qualitative results.** Visual representation of the frame selection of our method and the competitors in 2 new videos of each test set of the YouCook2 and COIN datasets. The Figure contains success cases, outlined in green, when our method accelerates and decelerates in the right segments, and cases in which our method fails, outlined in red, by accelerating ground-truth video clips or not decelerating clips belonging to the ground-truth.

validation set. As expected, the BoT approach is the slowest one since it includes a preprocessing step with dense captioning, saliency, and optical flow extraction followed by a frame scoring and a frame selection that includes a parameter setup with Particle Swarm Optimization and the shortest path selection in a graph. SAS and SASv2

run about $3\times$ faster, mainly due to the Minimum Sparse Reconstruction strategy instead of a graph in the frame selection step, which takes about $1.4$ms per frame ($\sim 714$ FPS). However, these approaches still require preprocessing steps like running the YOLO detector and extracting the optical flow. We highlight that our approach does not run

any preprocessing step, taking about $14\times$ less time than BoT and $4.5\times$ less than SAS and SASv2, with VDAN+ processing at 75ms per frame while SAFFA takes about $0.58$ms per frame. With a processing time of $8.65$ms, FFNet runs faster than all the other methods. The more significant advantage of FFNet over the other approaches is that its feature extraction step works only in the visual branch, with the AlexNet taking only 8ms per frame while the agent takes about $0.65$ms, making it run in real-time ($\sim 115$ FPS).



\end{document}